\documentclass[conference]{IEEEtran}
\usepackage{cite}
\usepackage{amsmath,amssymb,amsfonts}
\usepackage{algorithmic}
\usepackage{graphicx}
\usepackage{textcomp}
\usepackage{xcolor}
\usepackage{subcaption}
\usepackage[hyphens]{url}

\usepackage{amsmath,amsfonts,bm}

\def\eqref#1{equation~\ref{#1}}

\def\1{\bm{1}}

\DeclareMathAlphabet{\mathsfit}{\encodingdefault}{\sfdefault}{m}{sl}
\SetMathAlphabet{\mathsfit}{bold}{\encodingdefault}{\sfdefault}{bx}{n}

\usepackage{hyperref}
\usepackage{url}
\usepackage{minted}
\usepackage{subcaption}
\usepackage{amsmath}

\def\BibTeX{{\rm B\kern-.05em{\sc i\kern-.025em b}\kern-.08em
    T\kern-.1667em\lower.7ex\hbox{E}\kern-.125emX}}

\title{Finer is Better (with the Right Scaling)}

\author{Clemens Schaefer and Gil Tabak \\
Google LLC, Mountain View, Ca\\
\texttt{\{cjsschaefer,tabakg\}@google.com} \\
}

\begin{document}
\maketitle
\thispagestyle{plain}
\pagestyle{plain}












\begingroup\renewcommand\thefootnote{}%
\footnotetext{Experiment code available at \url{https://github.com/clee1994/finer_is_better}}%
\addtocounter{footnote}{-1}%
\endgroup

\begin{abstract}

Microscaling is a critical technique for preserving the quality of Large Language Models (LLMs) quantized to ultra-low precision formats. Intuitively, finer block sizes should yield lower quantization error; however, a paradox recently identified by Fasoli et al. (2026) demonstrates that standard abs-max scaling can actually result in degraded model quality as block sizes shrink. In this work, we investigate the underlying mechanics of this phenomenon. We demonstrate that this degradation is not an inherent limitation of finer granularity, but is primarily driven by how elements in smaller blocks statistically cluster closer to their local block maximum, interacting poorly with the coarse subnormal E4M3 values used as scaling factors. Specifically, we show that i) preventing the scaling factor from underflowing to zero mitigates errors caused by extreme underflow, ii) targeted algorithmic interventions like the 4-over-6 methodology that give more flexibility to the choice of scaling factor resolve the paradox for larger values, and iii) a brute-force search establishes an optimal baseline, confirming that the theoretical Mean Squared Error (MSE) strictly improves with finer block sizes. Ultimately, our findings highlight a critical insight for hardware-software co-design: the block-size paradox is partially an artifact of naive scale selection. While using hierarchical scaling factors or wider formats like UE5M3 interchangeably resolves much of the quality loss, we found the 4-over-6 scale selection heuristic can even further improve quality, especially for very small block sizes. Consequently, maximizing the performance of next-generation ML accelerators will require treating silicon format specifications and software scaling algorithms as tightly coupled design choices.

\end{abstract}

\section{Background}

The ubiquitous deployment of Large Language Models (LLMs) across datacenter architectures and edge devices incurs substantial computational overhead and expands the memory footprint, fundamentally constraining system throughput via both arithmetic and memory bandwidth bottlenecks. To address this, hardware vendors have begun natively supporting ultra-low precision formats (four bits or fewer) to simultaneously increase compute throughput and reduce memory pressure, as seen in NVIDIA's Blackwell and AMD's MI-series accelerators.~\cite{nvidia2025nvfp4,amd2024instinct}.

To offset quality losses associated with low bit formats, microscaling has proven to be a useful technique. The central idea is to enable chunking of the inputs into small blocks with individual scaling factors (e.g., block sizes of 16 or 32). Using microscales effectively on accelerators usually requires built-in hardware support. In common microscaling formats, the scaling factors themselves are also represented using low-precision formats (e.g., 8-bit floating-point), reducing the memory footprint and additional hardware burden of the greater number of scales. Usually, this is most powerful for very narrow formats like FP4 (E2M1), which otherwise lack the necessary dynamic range to represent a wide variety of distributional values.

MX v1.0~\cite{OCPMicroscaling2023} uses power-of-two scaling factors. Given inputs $V$ for each block, it specifies as one option the largest power-of-two less than or equal to the maximum absolute value of the input entries divided by the largest representable value in the element data type ($M_{\text{max}}$). Mathematically, this block scale $S$ is calculated as:

\begin{align}
\label{equ:mx_scale}
S = 2^{\lfloor \log_2 \left( \frac{\max_{V_i \in V} |V_i|}{M_{\text{max}}} \right) \rfloor}
\end{align}

For MXFP4, the element format is E2M1, where $M_{\text{max}} = 6.0$. Constraining the scaling factors to powers of two enables a high degree of hardware efficiency. Designing other quantization recipes is an area of active research. For example, \cite{chhugani2026unveiling} suggest mapping the scaling factor to the range $[3.5, 7.0]$; because clipping in the range $[6.0, 7.0]$ is mathematically equivalent to rounding in the range $[3.0, 3.5]$, this intentional clipping prevents underflows for smaller entries, resulting in less quantization error compared to no clipping at all.

Alternative microscaling specifications explore using 8-bit floating-point representations for the block scaling factors to provide finer numerical control. For instance, recent vendor-specific designs~\cite{abecassis2025pretraining, nvidia2025nvfp4} utilize an E4M3 representation (with the sign bit unused) for the block scale of size 16. The inclusion of mantissa bits in the scaling factor helps preserve higher precision, especially for the largest entry in each block. 
In typical applications the abs-max scale may be rounded to the nearest value in the scaling factor format. Because the exponent precision of E4M3 is relatively limited, this block scale is often paired with a second (per-tensor) scaling factor (i.e., hierarchical scaling).

The allocation of exponent versus mantissa bits can significantly influence the optimal quantization recipe (the conversion of higher-precision values into quantized entries along with their associated block-level scaling factors). For example, using all 8 bits for the exponent (unsigned E8M0) lowers the precision in the representation of values but covers a vast range of magnitudes. Redistributing some bits to the mantissa (e.g., E4M3 or UE5M3) lends higher precision to the entries at the expense of reducing the dynamic range, which for some distributions may require an additional tensor-level scaling factor.

The recent manuscript~\cite{fasoli2026finer} shows that surprisingly, when following the typical abs-max methodology, there can be a significant degradation in quality as the block sizes become smaller in particular when using E4M3 scaling factors. This seems paradoxical because smaller block sizes are supposed to give finer-grained control. In the remainder of this paper, we set up experiments to explain why this happens, how it can be mitigated, and explain the implications for scale factor and format selection.

\section{Experimental Setup}

We start by analyzing the microscale behavior for a variety of quantization techniques and several possible block sizes (4, 8, 16, 32). Unless stated otherwise, the data is quantized using a scaling factor format of E4M3 and the data is stored in FP4 (OCP E2M1)~\cite{ocp_ofp8_2023}.

FP4 has a small set of possible values: $\pm\{0.0, 0.5, 1.0, 1.5, 2.0, 3.0, 4.0, 6.0\}$. While the values are close near zero, they become farther between 4.0 and 6.0. 

We use combinations of strategies for selecting the scaling factor (discussed below).

To define abs-max scaling we first compute the scaling factor:

\begin{align}
\label{equ:abs-max-scale}
S_{\text{abs-max}}({M}) =  \text{Round}_{\text{SF}}\left( \frac{\max_{v \in V} |v|}{M } \right)
\end{align}

where the function $\text{Round}_{\text{SF}}$  includes the rounding and clipping to the target scale format (e.g., $\text{E4M3}$) and $M$ is a configurable maximum value (e.g., in the standard abs max scaling for E2M1, this would be $6.0$). We assume round-to-nearest-even throughout.
Abs-max scaling is often the default heuristic chosen in industry and was the primary method evaluated by~\cite{fasoli2026finer}. As we demonstrate later, this specific choice is closely tied to the paradoxical behavior observed.

For a given scaling factor $S$, quantized entries are found by applying the scaling and rounding using a function $\text{quant}$: $v_i \mapsto \text{quant}\left(\frac{v_i}{S} \right)$; de-quantization simply multiplies by $S$.

Because the scaling factor format is also low-precision (FP8), extremely small inputs can result in a scaling factor that rounds to zero. Since this preserves no information about the inputs (e.g. all values are set to zero), our first proposed adjustment modifies a scaling factor $S$ to disallow it from collapsing to zero:

\begin{align}
\label{equ:pz-scale}
S_{\text{prevent-zero}}({M}) = \max\left( S({M}), \epsilon_{\text{SF}} \right)
\end{align}

where $\epsilon_{\text{SF}}$ is the smallest nonzero value of the scaling factor format (for example $\epsilon_{\text{SF}}=2^{-9}$ for E4M3).

Additionally, we provide comparisons against the setup used in the four-over-six paper~\cite{cook2025four}, which considers two possibilities for the scaling factor (biggest and second biggest representable value), enabling better usage of the format for blocks that benefit more from one part of the E2M1 range versus another. 
Conceptually, this algorithm evaluates quantization error using both 4.0 and 6.0 as maximum bounds. By allowing the scale to adapt, it prevents the bulk of the block's elements from being forced into the wide, lossy quantization gap between 4.0 and 6.0 in the FP4 grid.

\begin{align}
S_{\text{4-over-6}}({M}) = \operatorname*{argmin}_{M\in \{4, 6 \}}
\text{MSE}(v, S_\text{abs-max}(M))
\end{align}
where $\text{MSE}(v, S)$ evaluates the quantization mean-squared error of the inputs $v$ with a scaling factor $S$.

For our full-model evaluations we also use \emph{hierarchical scaling} -- a two-stage process where a per-tensor scale is first identified using the abs-max across the entire tensor. Then, local scaling factors are found after adjusting the inputs using the global scale. Mathematically it is analogous to the process used in~\cite{abecassis2025pretraining} (Appendix B) with appropriate adjustments for other formats.

Finally, we establish an upper bound on performance via a brute-force search over possible scaling factors to minimize the MSE.

\section{Results}

We structure our experiments into two phases:

First, we conduct a numerical study based on sampling a normal distribution across a range of standard deviations and block sizes comparable to those explored in~\cite{fasoli2026finer}. We sample $50,000 \times \text{block\_size}$ values for each standard deviation $\sigma$ from the normal distribution $\mathcal{N}(0,\sigma^2)$, where the $\text{block\_size}$ is chosen from $\{4, 8, 16, 32\}$ and $\sigma$ is evaluated across $151$ evenly spaced points in the interval $[0.0005, 0.05]$.

Second, we validate our findings by applying these scaling methods to the weight and activation quantization (MLP layers) of several large language models to observe their impact on downstream perplexity. In the full model experiments, we consider different combinations of prevent-zero-scales, 4-over-6, hierarchical scaling and using the UE5M3 scaling factor suggested in~\cite{fasoli2026finer}.

\subsection{Numerical Study}
\label{section:numerical-study}

\begin{figure*}[t]
    \centering
    \includegraphics[width=\textwidth]{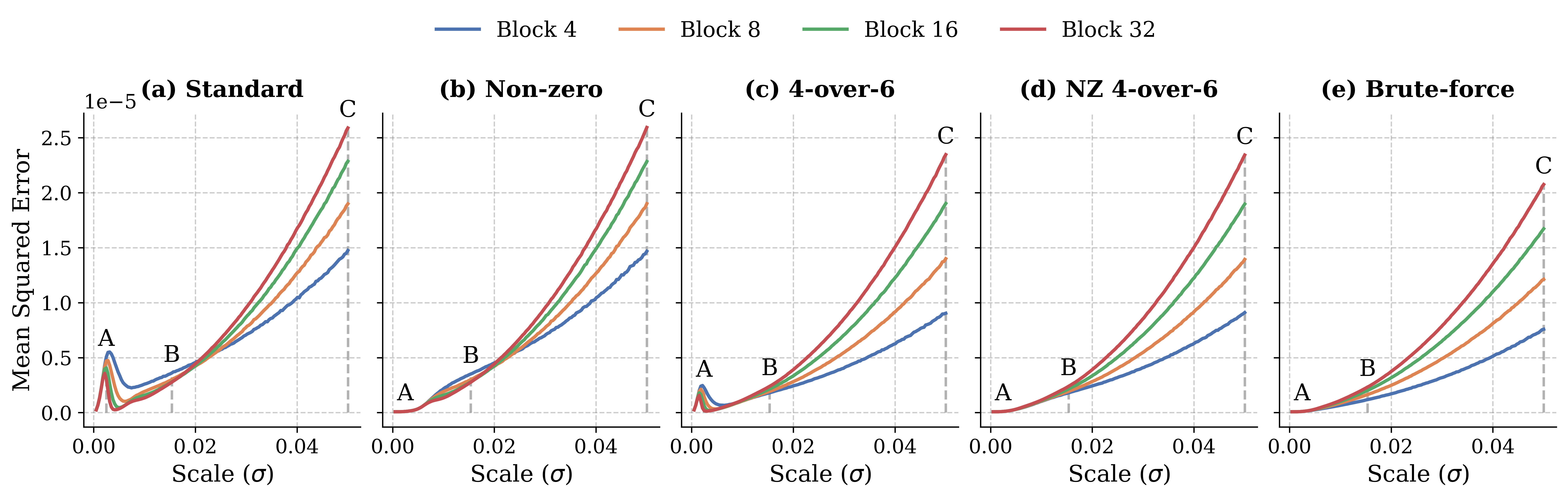}
    \vspace{-0.2cm} 
    \caption{Comparing MSE against distribution $\sigma$ behavior. \textbf{(a)} Standard max-abs scaling exhibits a visible bump in Region A. \textbf{(b)} Setting zero scales to the smallest representable value eliminates the Region A bump, but a reverse order of block sizes remains in Region B. \textbf{(c)} The 4-over-6 methodology retains the Region A bump but resolves Region B. \textbf{(d)} Shows the combination of prevent-zero and 4-over-6. \textbf{(e)} Brute-force search represents the optimal baseline.}
    \label{fig:mse_combined_horizontal}
\end{figure*}

Figure~\ref{fig:mse_combined_horizontal} illustrates the degradation of Mean Squared Error (MSE) when using standard abs-max scaling, comparing it to our proposed zero-prevention adjustment. We also evaluate the performance of the 4-over-6 scaling methodology against the theoretical minimum MSE achieved via a brute-force search.

The analysis reveals three distinct regions of behavior across the standard deviation ($\sigma$), denoted as A, B, and C. Under standard abs-max scaling, we observe the counter-intuitive phenomenon discovered in~\cite{fasoli2026finer}: smaller block sizes can paradoxically result in a higher MSE. 

In Region A (extremely small $\sigma$), the scaling factor frequently underflows to zero, resulting in an anomalous spike in MSE. By implementing the prevent-zero-scale method—forcing the scale to the smallest representable positive value—this localized degradation is completely eliminated. 

However, in Region B (slightly larger distribution standard deviations $\sigma$), the inversion of MSE relative to block size persists even with zero-prevention. This inversion is driven by a destructive interaction between the E4M3 scaling format and the statistics of small blocks. For values at this magnitude, the required E4M3 scaling factors fall into the subnormal range, making the scaling grid extremely coarse. Consequently, a block's true maximum value aligns poorly with the available scaling factors. When this coarse, poorly aligned scale is applied, it disproportionately impacts smaller blocks: because the maximum value in a block of 4 is statistically closer to the rest of the elements than in a block of 32, a much larger proportion of the elements are forced into the upper extremes of the FP4 format (near 6.0). Because the FP4 quantization grid is coarsest at these extremes, the result is massive clipping and rounding error. For more details about these mechanics see Appendix~\ref{appendix:appendix_histograms}.

This inversion vanishes by employing the 4-over-6 methodology. By adaptively evaluating both 4 and 6 as possible target scales, the 4-over-6 method gives the option to restrict values to the more uniform portion of the FP4 range, and to avoid strong clipping.

Finally, the brute-force search establishes an optimal baseline, demonstrating that when the ideal scaling factor is selected, the paradoxical inversion behavior disappears entirely. Across all points A, B, and C, the optimal MSE monotonically increases with larger block sizes, confirming that finer granularity fundamentally improves quantization quality when paired with an optimal scaling strategy.

At Region C (somewhat larger values) we never see the inversion--however, 4-over-6 and the brute force methods have significantly lower MSE than the abs-max based methods especially at smaller block sizes (explained similarly by the large potion of entries at the bins of magnitude 6).

\subsection{Downstream LLM Perplexity}

\begin{figure*}[t]
    \centering
    \includegraphics[width=\textwidth]{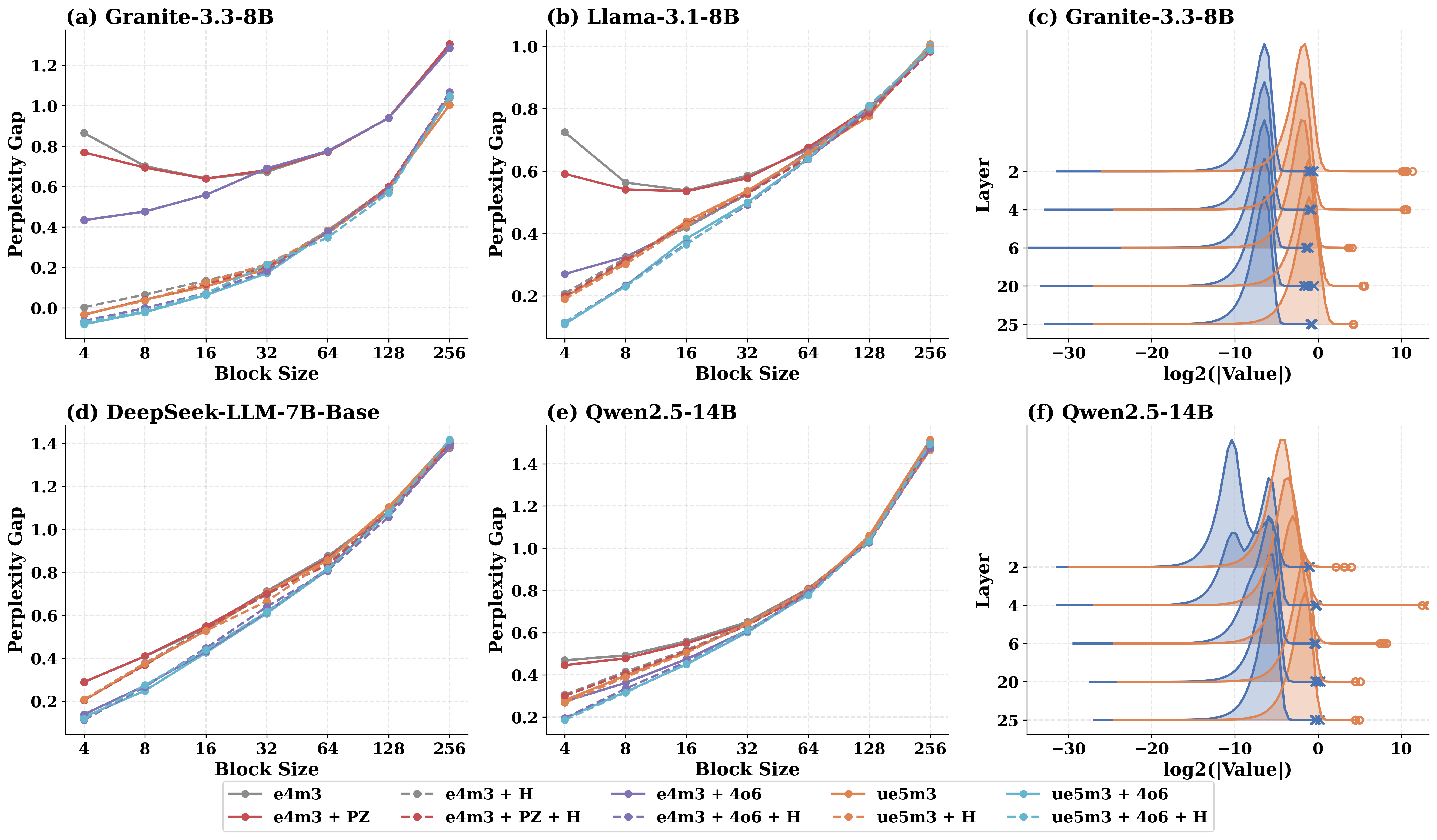}
    \caption{\textbf{(a,b,d,e)} show the perplexity gap of the Granite-3.3-8B, Llama-3.1-8B, DeepSeek-7B, and Qwen2.5-14B models with various FP4 formats compared to a unquantized baseline. Note that only Granite and Llama demonstrate the paradoxical increased perplexity gap at lower block sizes, meanwhile DeepSeek and Qwen improve the perplexity with smaller block sizes. Techniques such as hierarchical scales (H), four over six (\texttt{4o6}) ~\cite{cook2025four}, and more exponent bits (\texttt{ue5m3}) fully eliminate the ill behaved Granite and Llama at smaller block sizes. \textbf{(c,f)} show representative activation and weight distribution of Granite and Qwen for layers 2, 4, 6, 20, and 25, where blue represents the weights and orange the activations.}
    \label{fig:final_figure}
\end{figure*}

To confirm that the theoretical MSE improvements translate to practical applications, we evaluated the perplexity gap of four distinct models—Granite-3.3-8B, Llama-3.1-8B, DeepSeek-LLM-7B-Base, and Qwen2.5-14B—across multiple block sizes, comparing various FP4 quantization formats against an unquantized baseline (Figure~\ref{fig:final_figure}).

A key observation from these evaluations is a stark dichotomy in model behavior under abs-max quantization. DeepSeek and Qwen behave intuitively: as the block size decreases (offering finer-grained scaling), the perplexity gap narrows, indicating improved quality. Conversely, Granite and Llama clearly exhibit the paradoxical degradation identified in our numerical study and~\cite{fasoli2026finer}; for these models, standard E4M3 scaling results in a sharply increased perplexity gap at smaller block sizes, most notably at block sizes 4 and 8. 

By systematically evaluating different scaling configurations, we demonstrate how this non-intuitive behavior can be successfully mitigated. Isolating our proposed prevent-zero-scale (PZ) adjustment reveals noticeable but localized improvements; for instance, at the problematic block size of 4, the PZ adjustment reduces Granite's perplexity gap from 0.865 to 0.769, and Llama's from 0.725 to 0.591. However, we find that the 4-over-6 methodology~\cite{cook2025four} can have significantly greater impact on reducing the perplexity gap. Applying 4-over-6 alone reduces the perplexity gap by more than half for the ill-behaved models at the finest granularity, dropping Llama to 0.270 and Granite to 0.434 at block size 4. While the PZ adjustment does not resolve the U-shaped curve seen for Granite and Llama, 4-over-6 does.

Additionally, our results help discern the impact of hierarchical scaling and the 4-over-6 method and compare them to changing the scale format to UE5M3. Specifically, using E4M3 with a hierarchical scale (\texttt{e4m3 + H}) yields comparable results to using UE5M3, and this change is synergetic with including 4-over-6 scale selection--that is, using E4M3 with both hierarchical scales and 4-over-6 (\texttt{e4m3 + 4o6 + H}) yields better results, and is comparable to using UE5M3 with 4-over-6 (\texttt{ue5m3 + 4o6}). Finally, incorporating hierarchical scaling with UE5M3 regardless of whether using 4-over-6 does not make an impact beyond simply UE5M3 (\texttt{ue5m3} vs. \texttt{ue5m3 + H} and \texttt{ue5m3 + 4o6} vs. \texttt{ue5m3 + 4o6 + H}). These results suggest that using either E4M3 with hierarchical scaling or UE5M3 are sufficient for covering the important range of the data, \textbf{although there is an additional separate impact from the scale selection technique}.

Finally, we examine the raw model weight and activation distributions to understand this divergence in model behavior. Figure~\ref{fig:final_figure} (c, f) highlights these distributions for Granite and Qwen. Despite a large portion of Qwen's weights having small magnitudes, it is relatively robust to E4M3 scaling variants. To understand why Granite degrades while Qwen improves, we conducted an ablation study masking specific magnitude intervals to zero (see Appendix~\ref{appendix:weights}, Fig.~\ref{fig:zones_zero}). We found that Qwen's performance relies heavily on larger magnitude weights, shielding it from the coarse E4M3 subnormal gap. Conversely, Granite relies heavily on smaller values that are effectively destroyed by coarse upper bins. This ablation acts as a pre-quantization diagnostic: rather than relying solely on raw tensor distributions, masking specific magnitude intervals can successfully predict a model's sensitivity to specific entries.

Adopting \texttt{ue5m3} instead inherently reduces this degradation by offering a wider absolute range—often circumventing the need for an additional tensor-level scale.

\textbf{Hardware Tradeoffs:} Vendors must weigh the physical silicon area costs of supporting a 5-bit exponent in hardware ALUs (UE5M3) against the memory bandwidth and latency overheads of fetching a secondary, tensor-level scale from memory (hierarchical scaling). In our experiments, both approaches yield similar quality improvements, though UE5M3 offers greater general flexibility. Furthermore, while the 4-over-6 algorithm significantly improves quality, it incurs substantial computational overhead by requiring two quantizations and MSE calculations per block. This is reasonable for inference weights (which are quantized offline), but the latency overhead may be prohibitive for dynamic activation quantization.

While our evaluation spans models from 7B to 14B parameters—a substantial and widely deployed scale—this places our study right at the critical threshold of outlier emergence. Prior literature demonstrates that extreme activation outliers systematically appear as models cross the 6.7B and 13B parameter boundaries, escalating in larger architectures~\cite{dettmers2022gpt3}. Because these massive activations grow with the model size~\cite{sun2024massive}, the heavy-tailed distributions that disrupt the FP4 quantization grid are likely to become even more pronounced in 70B+ regimes. Additionally, recent work has uncovered intra-level sparsity~\cite{park2026uncovering}, which may also interact with low-precision quantization recipes. Verifying how scaling heuristics hold up against these aggressive outliers and sparse regimes at massive scales remains an important direction for future research.

\section{Conclusion}

This work investigates the paradoxical degradation of quantization quality at finer microscaling block sizes under standard abs-max scaling. Through numerical analysis and downstream LLM evaluation, we demonstrate that this anomaly is not an inherent limitation of finer granularity, but rather a consequence of how elements in smaller blocks statistically cluster closer to the per-block maximum, interacting harmfully with the coarse upper quantization bins of the FP4 (E2M1) element format. 

While enforcing a strictly positive minimum scale (prevent-zero) acts as a computationally lightweight mitigation for extreme underflow, it is insufficient to resolve the structural paradox for slightly larger values. Resolving the paradox on real, deployed models requires targeted algorithmic interventions, such as the 4-over-6 strategy, to prevent elements from being forced into wide quantization gaps. 

Crucially, our brute-force baseline reveals that the block-size paradox is entirely an artifact of the scale-selection heuristic, not a limitation of the FP4 format itself. This highlights a critical gap in current hardware-software co-design: while vendors are rapidly adopting ultra-low precision formats, the standard arithmetic heuristics (like abs-max) used to map values into these formats leave significant theoretical performance on the table. Unlocking the true potential of microscaling will require moving beyond naive maximums and developing silicon-friendly heuristics that approximate optimal scale selection without incurring prohibitive computational latency.


\section*{Acknowledgment}
The authors would like to thank Denali Molitor for careful review of the manuscript.

\bibliographystyle{IEEEtran}
\bibliography{iclr2026_conference}

@article{fasoli2026finer,
  title={Is Finer Better? The Limits of Microscaling Formats in Large Language Models},
  author={Fasoli, Andrea and Kar, Monodeep and Liu, Chi-Chun and Venkataramani, Swagath and Srinivasan, Viji and Chang, Leland and Wang, Naigang},
  journal={arXiv preprint arXiv:2601.19026},
  year={2026}
}

@article{cook2025four,
  title={Four Over Six: More Accurate NVFP4 Quantization with Adaptive Block Scaling},
  author={Cook, Jack and Guo, Junxian and Xiao, Guangxuan and Lin, Yujun and Han, Song},
  journal={arXiv preprint arXiv:2512.02010},
  year={2025}
}

@techreport{OCPMicroscaling2023,
  author      = {Bita Darvish Rouhani and Nitin Garegrat and Tom Savell and Ankit More and Kyung-Nam Han and Ritchie Zhao and Mathew Hall and Jasmine Klar and Eric Chung and Yuan Yu and Michael Schulte and Ralph Wittig and Ian Bratt and Nigel Stephens and Jelena Milanovic and John Brothers and Pradeep Dubey and Marius Cornea and Alexander Heinecke and Andres Rodriguez and Martin Langhammer and Summer Deng and Maxim Naumov and Paulius Micikevicius and Michael Siu and Colin Verrilli},
  title       = {Open Compute Project OCP Microscaling Formats (MX) Specification},
  institution = {Open Compute Project},
  year        = {2023},
  month       = {September},
  version     = {1.0},
  url         = {https://www.opencompute.org/documents/ocp-microscaling-formats-mx-v1-0-spec-final-pdf}
}

@article{abecassis2025pretraining,
  title={Pretraining large language models with nvfp4},
  author={Abecassis, Felix and Agrusa, Anjulie and Ahn, Dong and Alben, Jonah and Alborghetti, Stefania and Andersch, Michael and Arayandi, Sivakumar and Bjorlin, Alexis and Blakeman, Aaron and Briones, Evan and others},
  journal={arXiv preprint arXiv:2509.25149},
  year={2025}
}

@article{chhugani2026unveiling,
  title={Unveiling the Potential of Quantization with MXFP4: Strategies for Quantization Error Reduction},
  author={Chhugani, Jatin and Jeong, Geonhwa and Su, Bor-Yiing and Pan, Yunjie and Yang, Hanmei and Ankit, Aayush and Yu, Jiecao and Deng, Summer and Chen, Yunqing and Satish, Nadathur and others},
  journal={arXiv preprint arXiv:2603.08713},
  year={2026}
}

@techreport{ocp_ofp8_2023,
    title = {{OCP 8-bit Floating Point Specification (OFP8) Revision 1.0}},
    author = {Micikevicius, Paulius and Oberman, Stuart and Dubey, Pradeep and Cornea, Marius and Rodriguez, Andres and Bratt, Ian and others},
    institution = {Open Compute Project},
    year = {2023},
    month = {December},
    url = {https://www.opencompute.org/documents/ocp-8-bit-floating-point-specification-ofp8-revision-1-0-2023-12-01-pdf-1}
}

@misc{nvidia2025nvfp4,
  title={Introducing {NVFP4} for Efficient and Accurate Low-Precision Inference},
  author={{NVIDIA Corporation}},
  howpublished={\url{https://developer.nvidia.com/blog/introducing-nvfp4-for-efficient-and-accurate-low-precision-inference/}},
  year={2025},
  note={Accessed: 2026-04-23}
}

@misc{amd2024instinct,
  title={{AMD Instinct MI300} Series Accelerators and the Era of Adaptive Microscaling},
  author={{Advanced Micro Devices, Inc.}},
  howpublished={\url{https://www.amd.com/en/products/accelerators/instinct.html}},
  year={2024},
  note={Accessed: 2026-04-23}
}

@article{park2026uncovering,
  title={Uncovering Intra-expert Activation Sparsity for Efficient Mixture-of-Expert Model Execution},
  author={Park, Jongseok and Kim, Sunga and Gu, Zhenyu and Stoica, Ion and Cheung, Alvin},
  journal={arXiv preprint arXiv:2605.08575},
  year={2026}
}

@article{sun2024massive,
  title={Massive activations in large language models},
  author={Sun, Mingjie and Chen, Xinlei and Kolter, J Zico and Liu, Zhuang},
  journal={arXiv preprint arXiv:2402.17762},
  year={2024}
}

@article{dettmers2022gpt3,
  title={Gpt3. int8 (): 8-bit matrix multiplication for transformers at scale},
  author={Dettmers, Tim and Lewis, Mike and Belkada, Younes and Zettlemoyer, Luke},
  journal={Advances in neural information processing systems},
  volume={35},
  pages={30318--30332},
  year={2022}
}

\appendix
\section{Appendix}
\label{appendix:appendix_histograms}

\subsection{Entry and Scale Histograms for Normal Distributions}

The following histograms illustrate the absolute quantized values and the corresponding scaling factors selected by each methodology across the three critical standard deviations ($\sigma$) identified as Regions A, B, and C in the numerical study discussed in Section~\ref{section:numerical-study} and shown in Figure~\ref{fig:mse_combined_horizontal}. Within each figure, we show the histograms for the entries (left side) along with their corresponding scaling factors (right side). The color and row within each figure represent the four choices of block size. The shaded-out red region represents values not representable using the FP4 format.

These diagrams can help understand the numerical behavior under quantization using different recipes by inspecting the utilization of each bin of the entry and scale formats.

\begin{figure*}[htbp]
    \centering
    \begin{subfigure}[b]{0.32\textwidth}
        \centering
        \includegraphics[width=\textwidth]{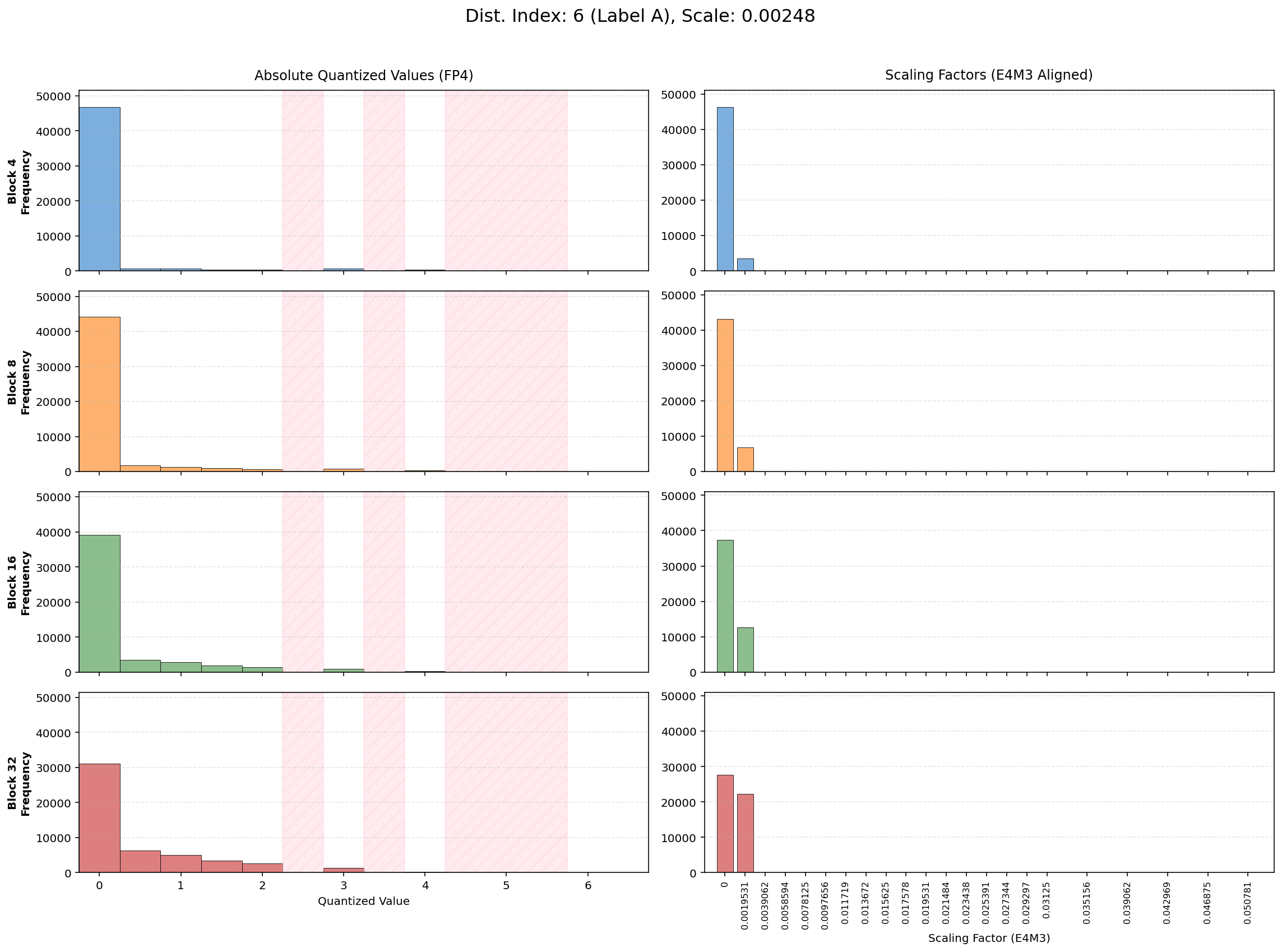}
        \caption{Region A}
        \label{fig:hist-abs-max-a}
    \end{subfigure}
    \hfill
    \begin{subfigure}[b]{0.32\textwidth}
        \centering
        \includegraphics[width=\textwidth]{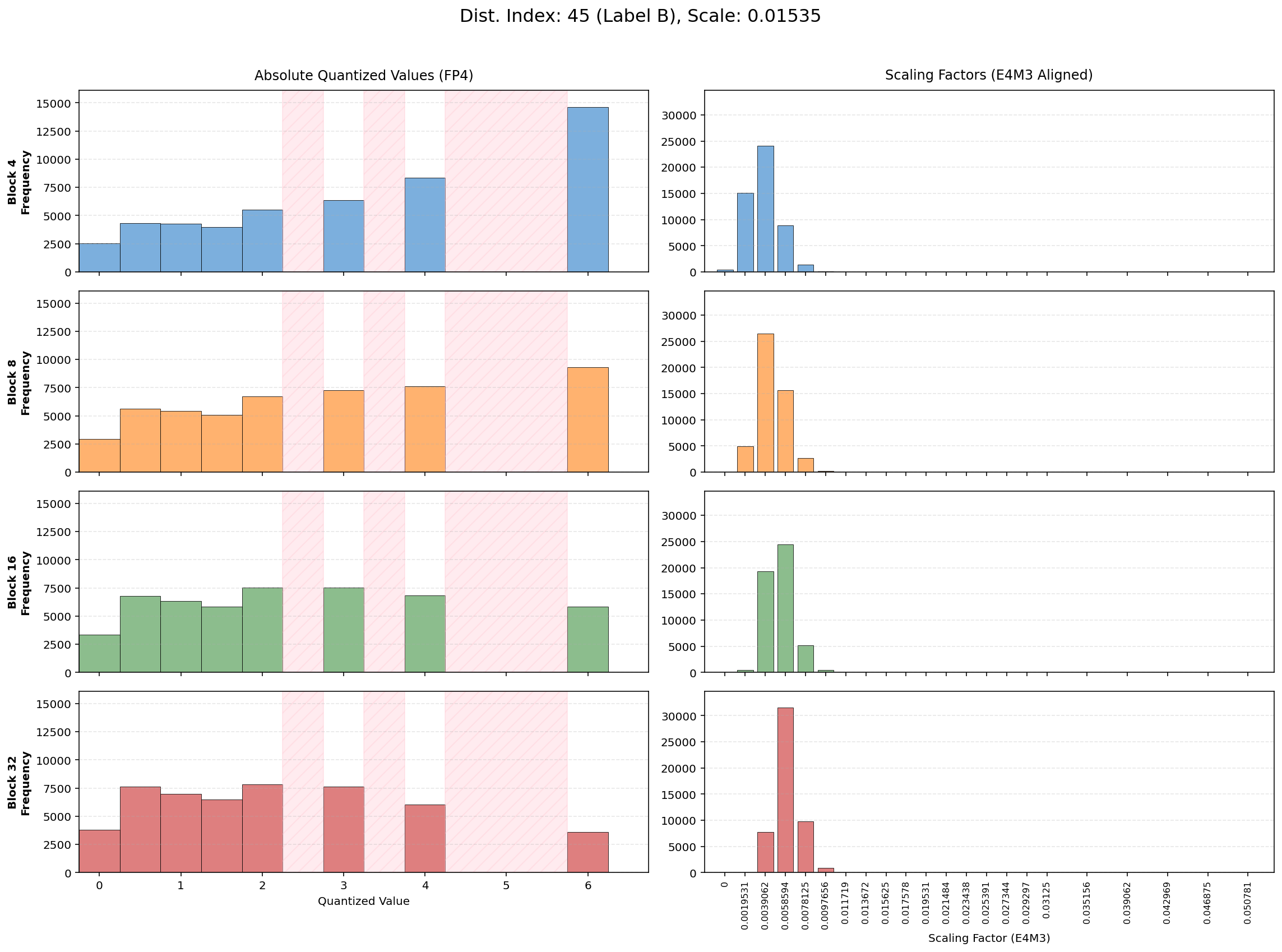}
        \caption{Region B}
        \label{fig:hist-abs-max-b}
    \end{subfigure}
    \hfill
    \begin{subfigure}[b]{0.32\textwidth}
        \centering
        \includegraphics[width=\textwidth]{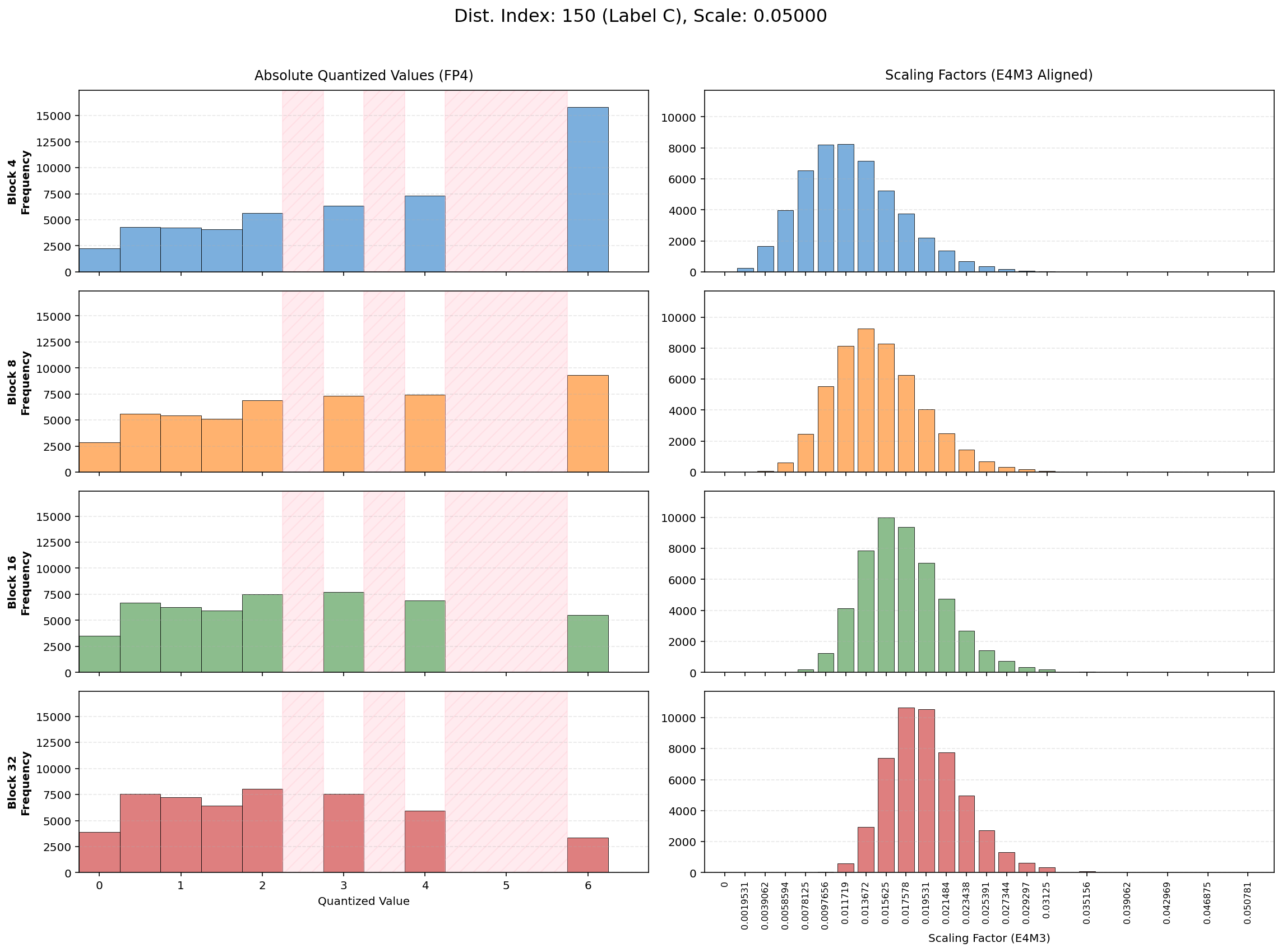}
        \caption{Region C}
        \label{fig:hist-abs-max-c}
    \end{subfigure}
    \vspace{-0.2cm}
    \caption{Standard Abs-Max Scaling. \textbf{(a) Region A:} At extremely small $\sigma$, the standard algorithm frequently selects a scaling factor of zero, obliterating element information and causing a sharp MSE spike. \textbf{(b) Regions B and C:} At these ranges, we find smaller blocks have more entries at the bin of magnitude 6. At Region B the scales are relatively concentrated at only a few bins, while at Region C they are spread out across a wider range.}
    \label{fig:hist-abs-max-combined}
\end{figure*}

\begin{figure*}[htbp]
    \centering
    \begin{subfigure}[b]{0.32\textwidth}
        \centering
        \includegraphics[width=\textwidth]{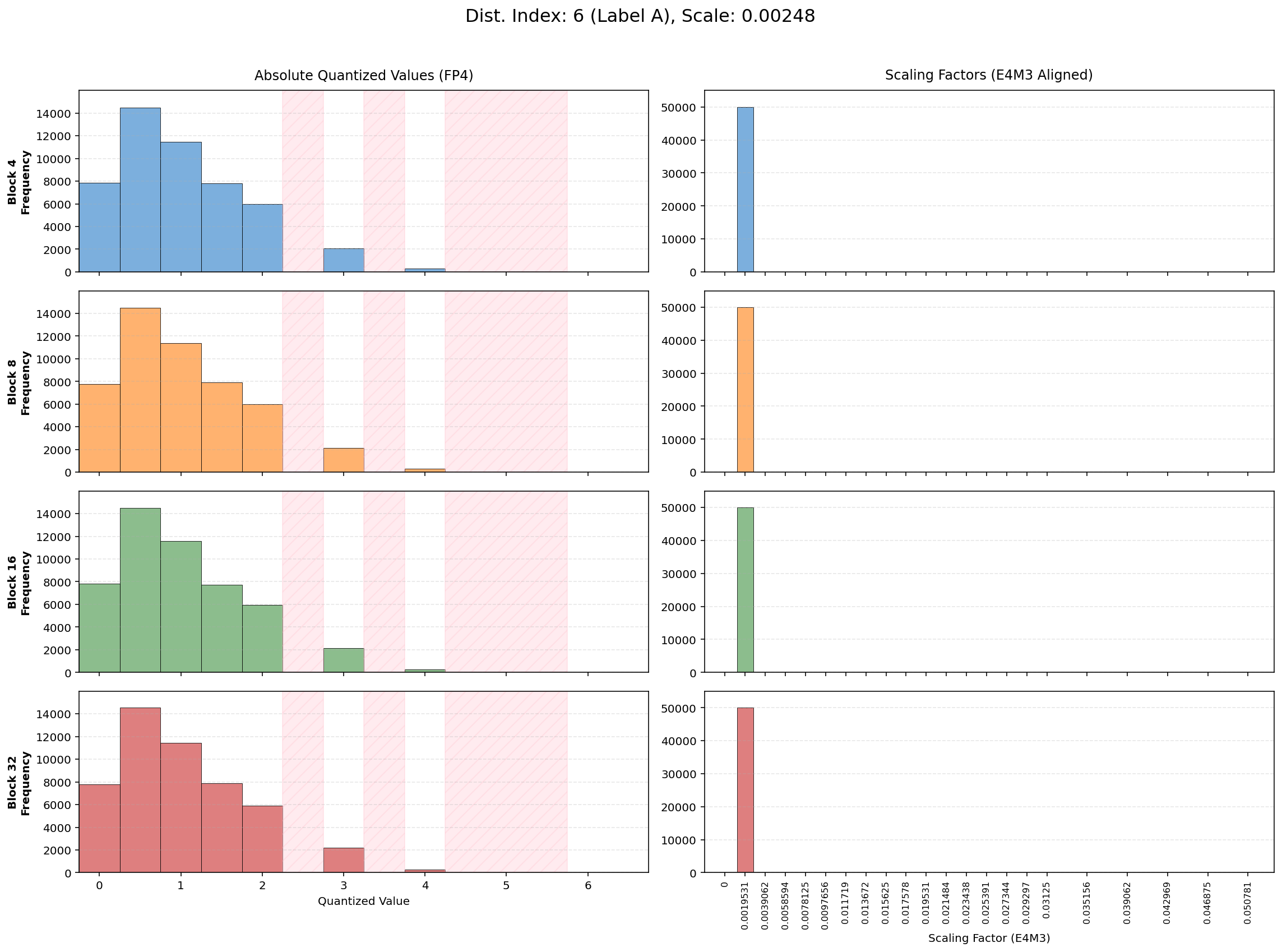}
        \caption{Region A}
        \label{fig:hist-prevent-zero-a}
    \end{subfigure}
    \hfill
    \begin{subfigure}[b]{0.32\textwidth}
        \centering
        \includegraphics[width=\textwidth]{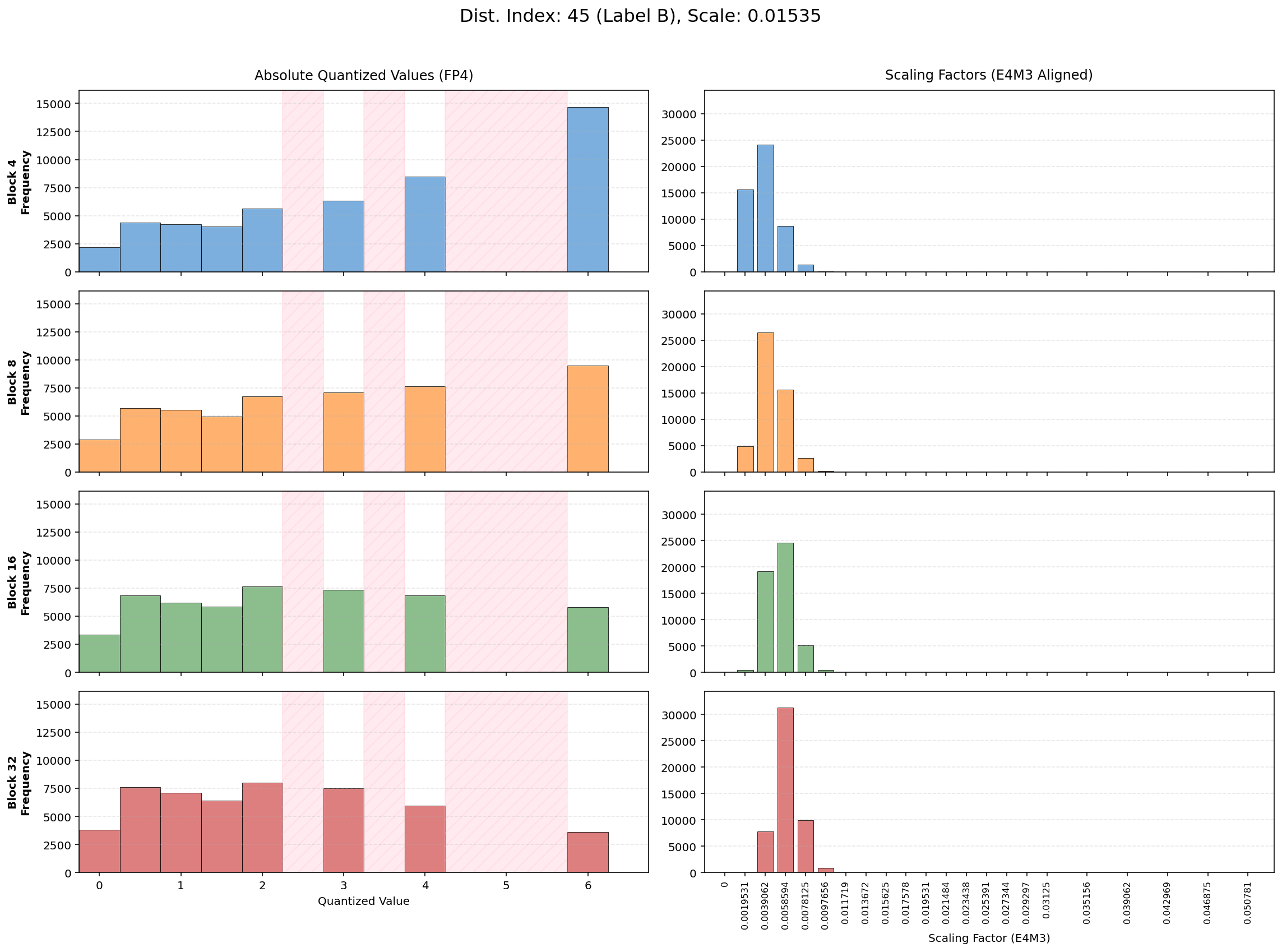}
        \caption{Region B}
        \label{fig:hist-prevent-zero-b}
    \end{subfigure}
    \hfill
    \begin{subfigure}[b]{0.32\textwidth}
        \centering
        \includegraphics[width=\textwidth]{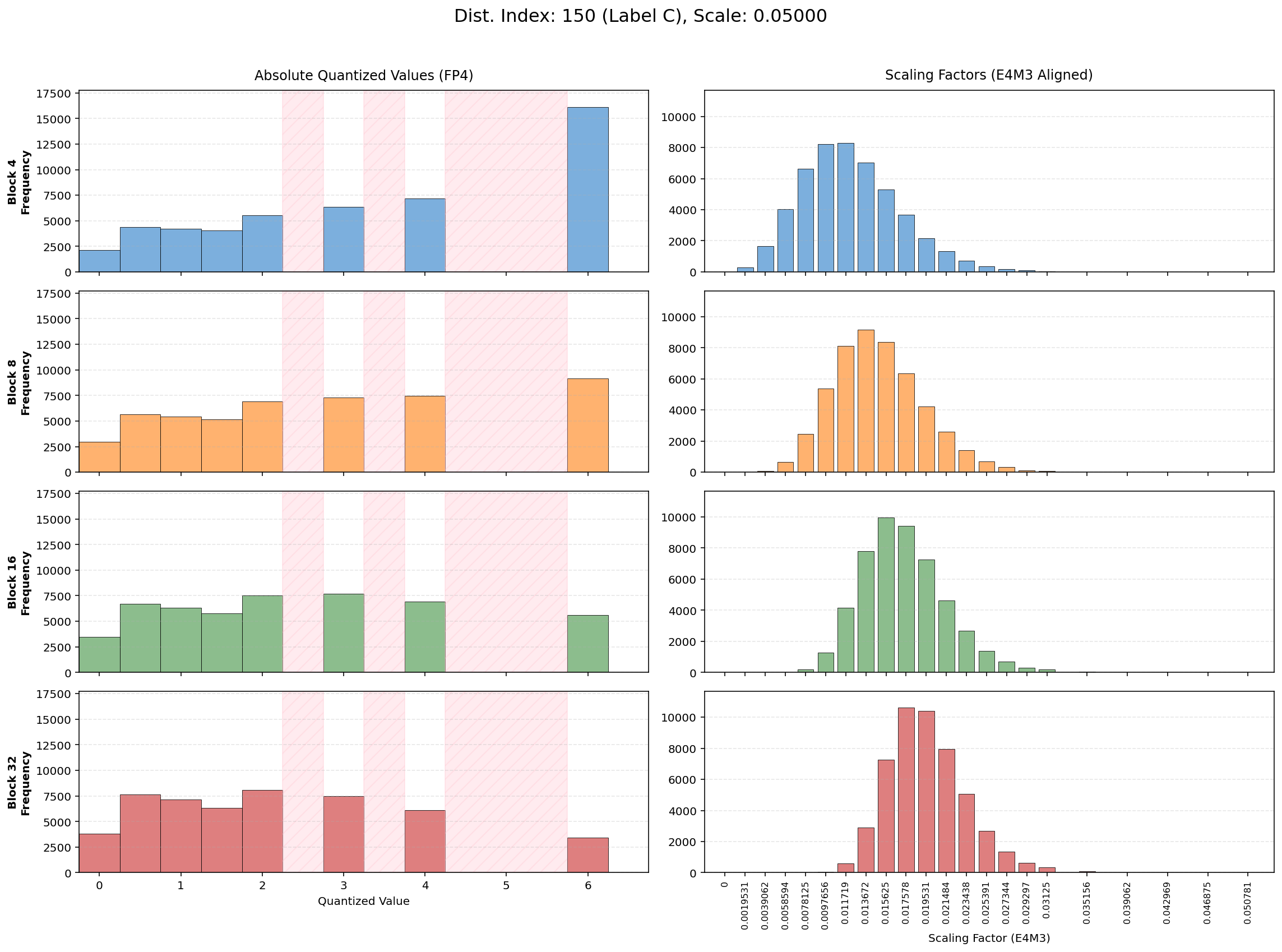}
        \caption{Region C}
        \label{fig:hist-prevent-zero-c}
    \end{subfigure}
    \vspace{-0.2cm}
    \caption{Prevent-Zero Adjustment. \textbf{(a) Region A:} Restricting the scaling factor to the smallest positive representable value avoids a collapse to zero, preserving information and eliminating the localized MSE spike. \textbf{(b) Region B:} Elements remain pushed into the coarse quantization gap between 4 and 6, visually confirming that preventing zero scales does not resolve the structural element-level anomaly. \textbf{(c) Region C:} Behavior matches the standard abs-max methodology.}
    \label{fig:hist-prevent-zero-combined}
\end{figure*}

\begin{figure*}[htbp]
    \centering
    \begin{subfigure}[b]{0.32\textwidth}
        \centering
        \includegraphics[width=\textwidth]{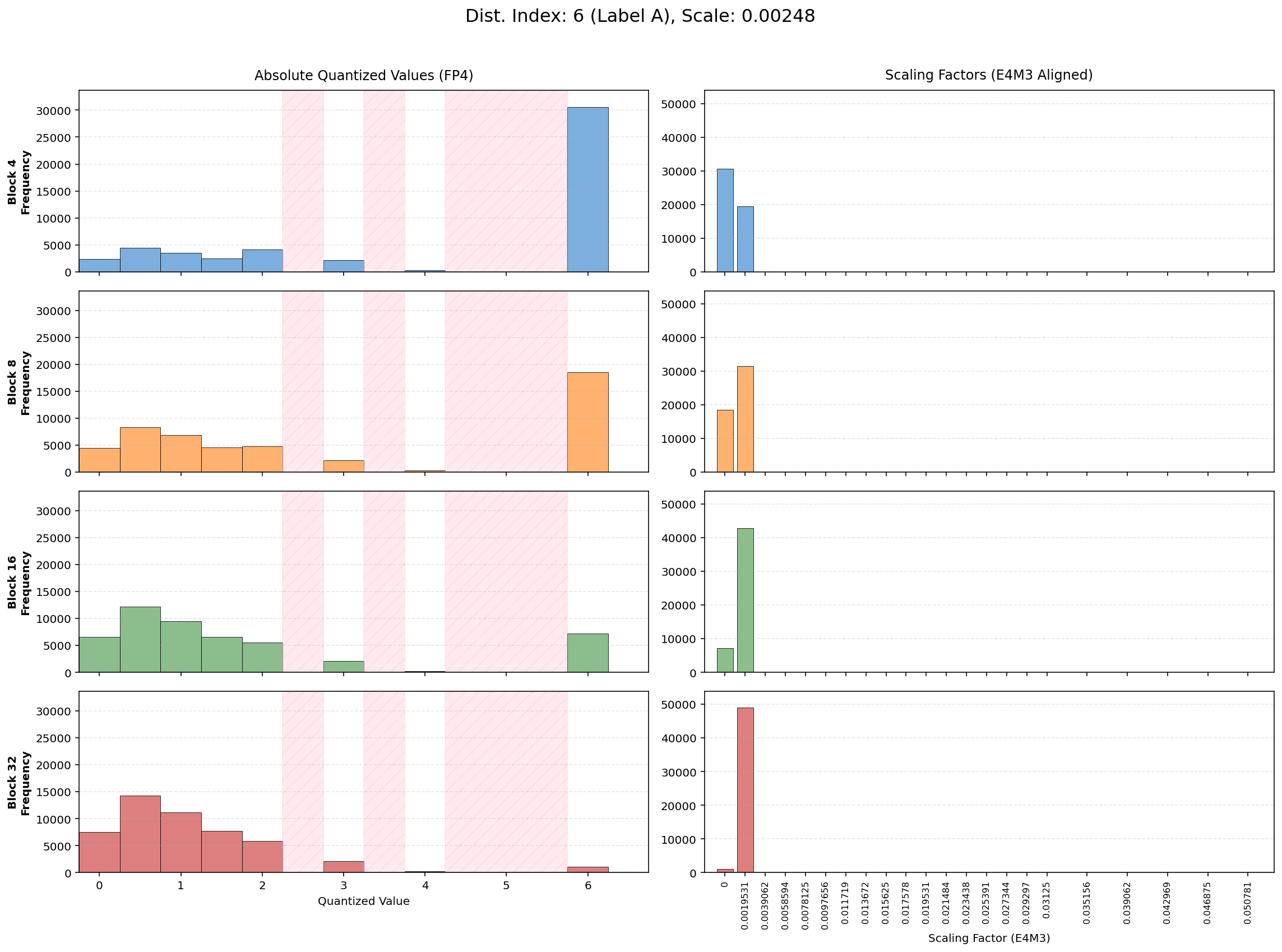}
        \caption{Region A}
        \label{fig:hist-4-over-6-a}
    \end{subfigure}
    \hfill
    \begin{subfigure}[b]{0.32\textwidth}
        \centering
        \includegraphics[width=\textwidth]{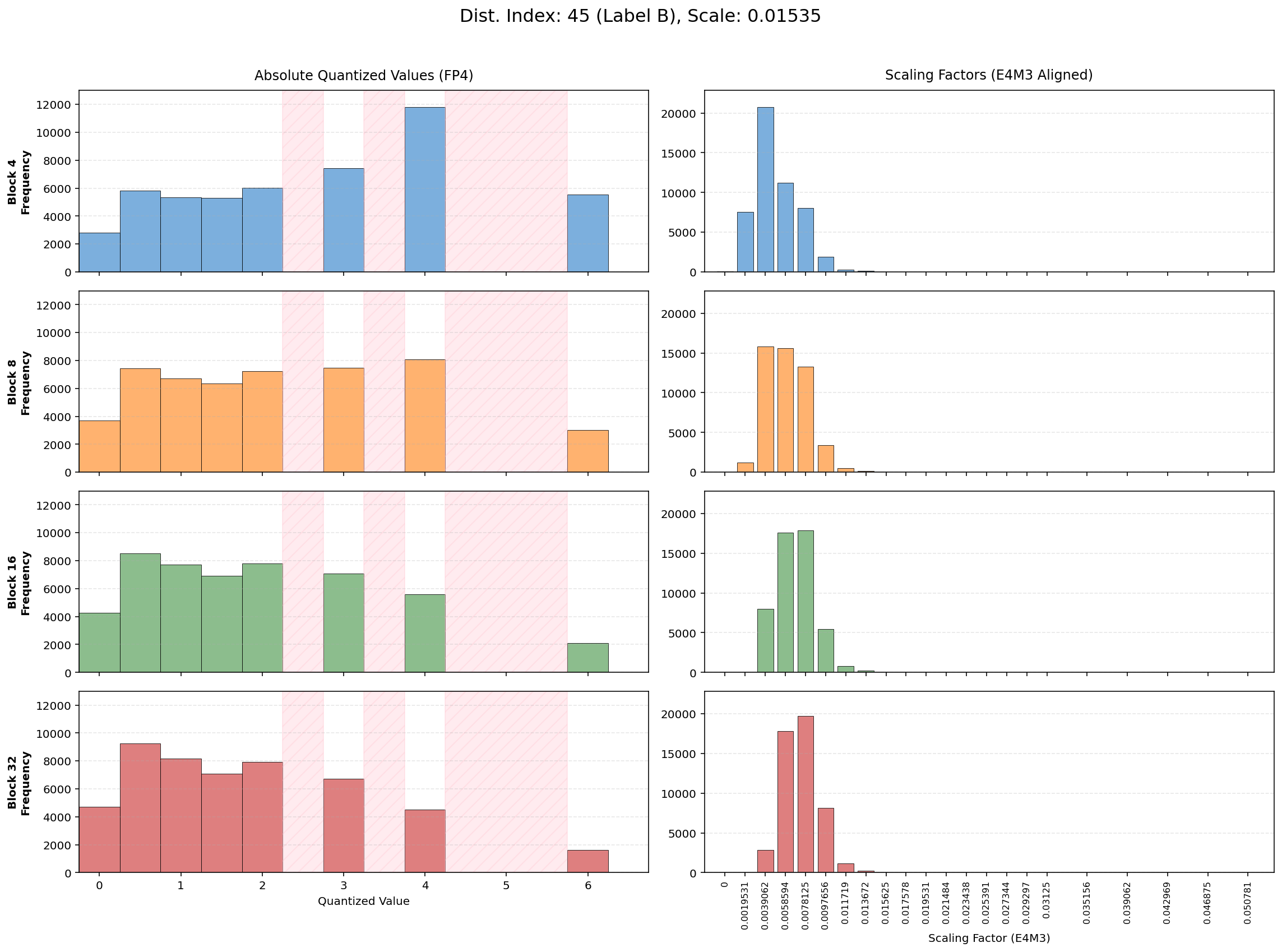}
        \caption{Region B}
        \label{fig:hist-4-over-6-b}
    \end{subfigure}
    \hfill
    \begin{subfigure}[b]{0.32\textwidth}
        \centering
        \includegraphics[width=\textwidth]{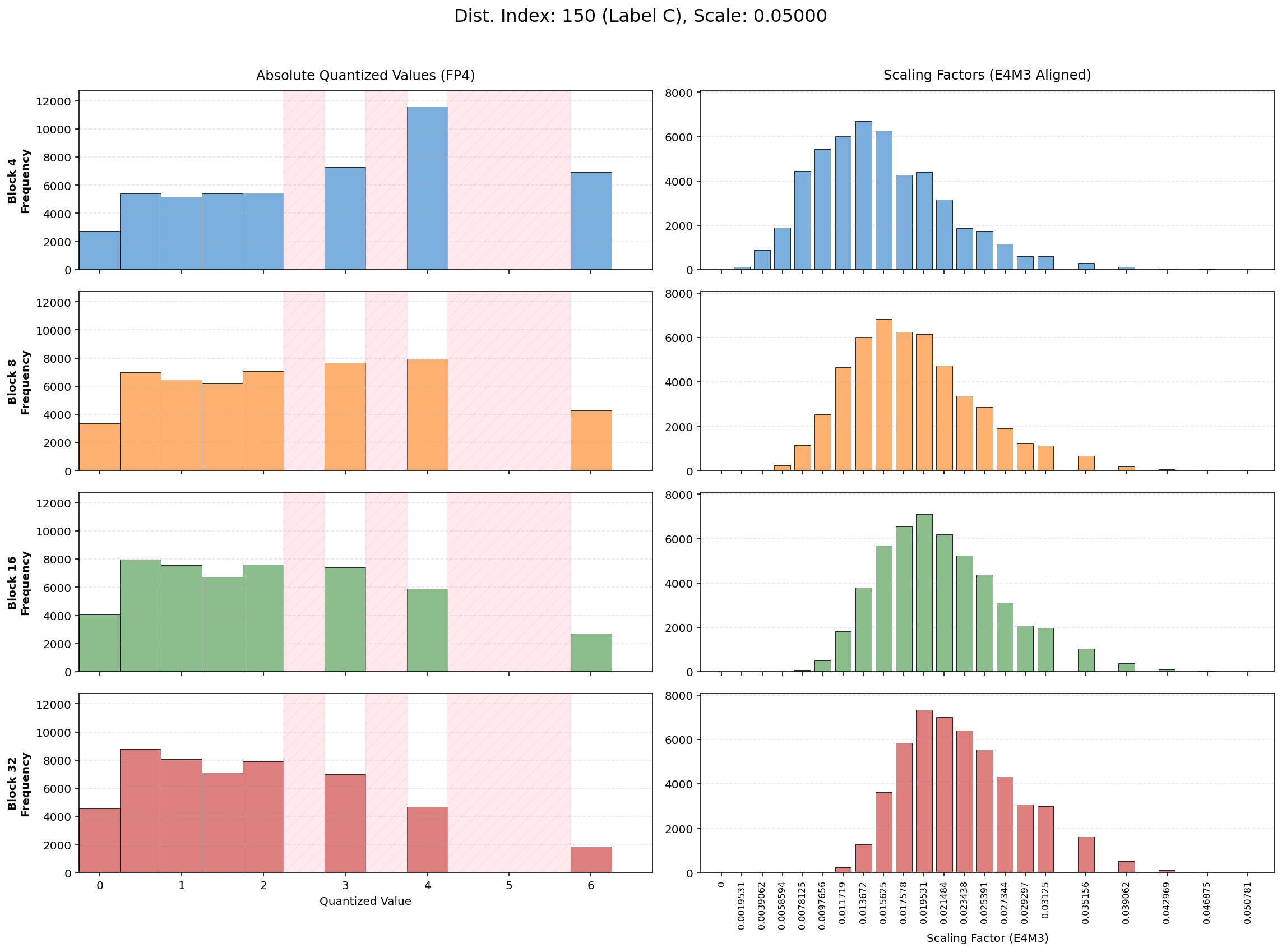}
        \caption{Region C}
        \label{fig:hist-4-over-6-c}
    \end{subfigure}
    \vspace{-0.2cm}
    \caption{4-over-6 Methodology. \textbf{(a) Region A:} While providing better overall bin utilization, the scaling factors can still struggle with extreme underflow in this tiny magnitude regime. \textbf{(b) Region B:} By adaptively allowing a maximum target scale of 4, the algorithm successfully avoids forcing elements into the massive upper quantization gap of FP4, resolving the scaling inversion paradox. \textbf{(c) Region C:} Similar to Region B, with scaling factors spread across a wider set of bins. In both Regions B and C, while there are fewer entries aggregated at the bin of magnitude 6 for small blocks, there are more at the bin of magnitude 4.}
    \label{fig:hist-4-over-6-combined}
\end{figure*}

\begin{figure*}[htbp]
    \centering
    \begin{subfigure}[b]{0.32\textwidth}
        \centering
        \includegraphics[width=\textwidth]{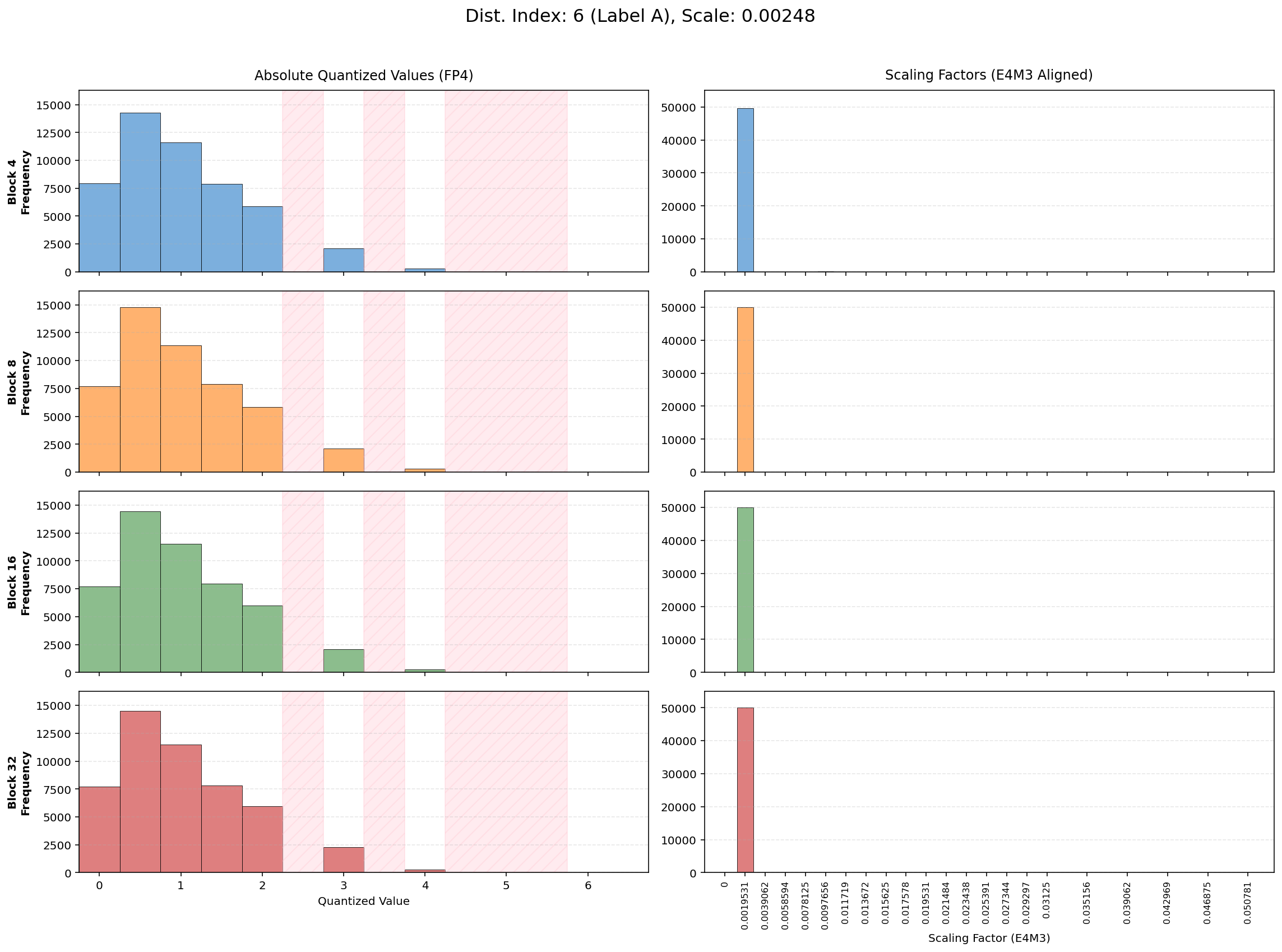}
        \caption{Region A}
        \label{fig:hist-brute-force-a}
    \end{subfigure}
    \hfill
    \begin{subfigure}[b]{0.32\textwidth}
        \centering
        \includegraphics[width=\textwidth]{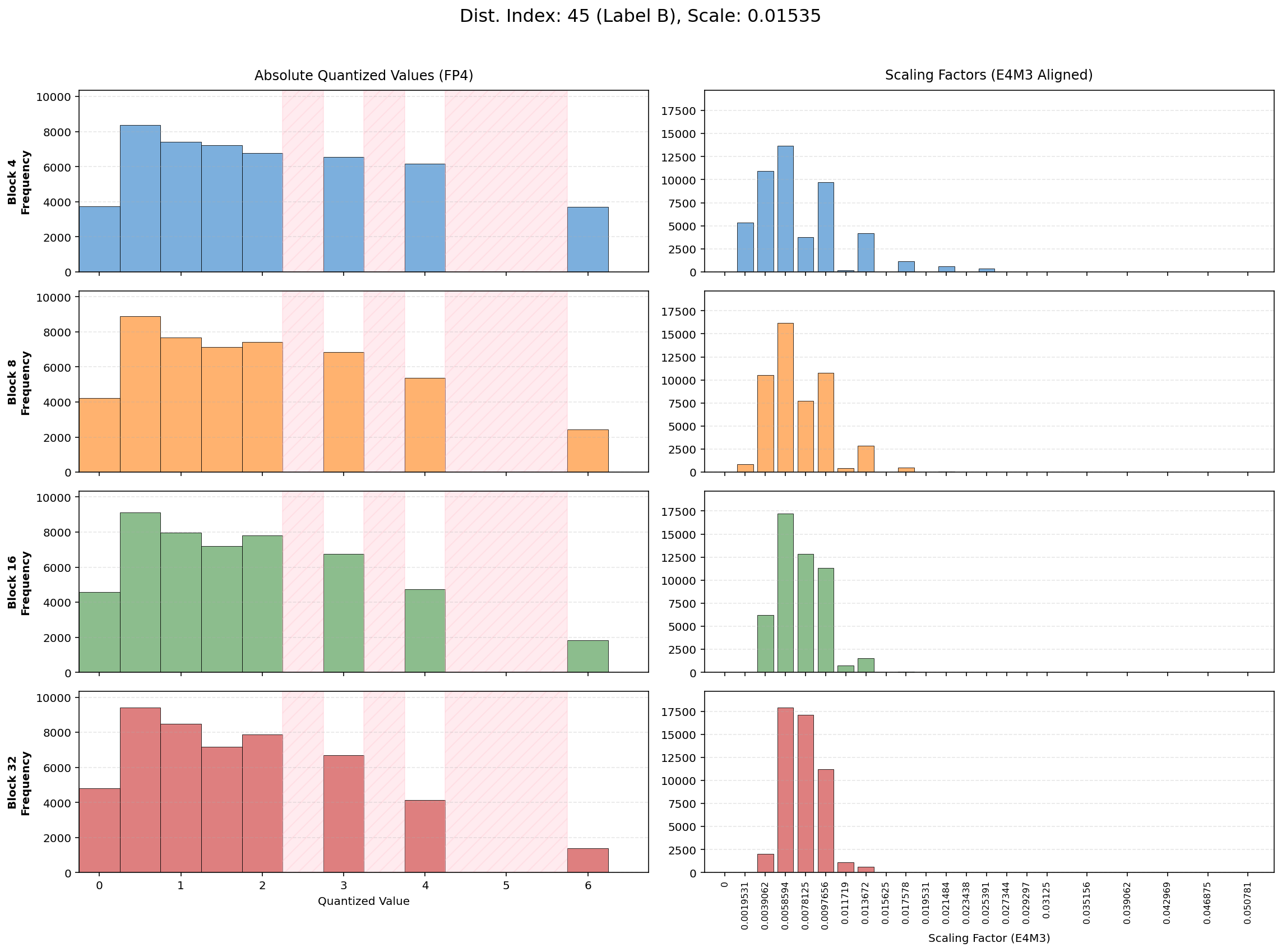}
        \caption{Region B}
        \label{fig:hist-brute-force-b}
    \end{subfigure}
    \hfill
    \begin{subfigure}[b]{0.32\textwidth}
        \centering
        \includegraphics[width=\textwidth]{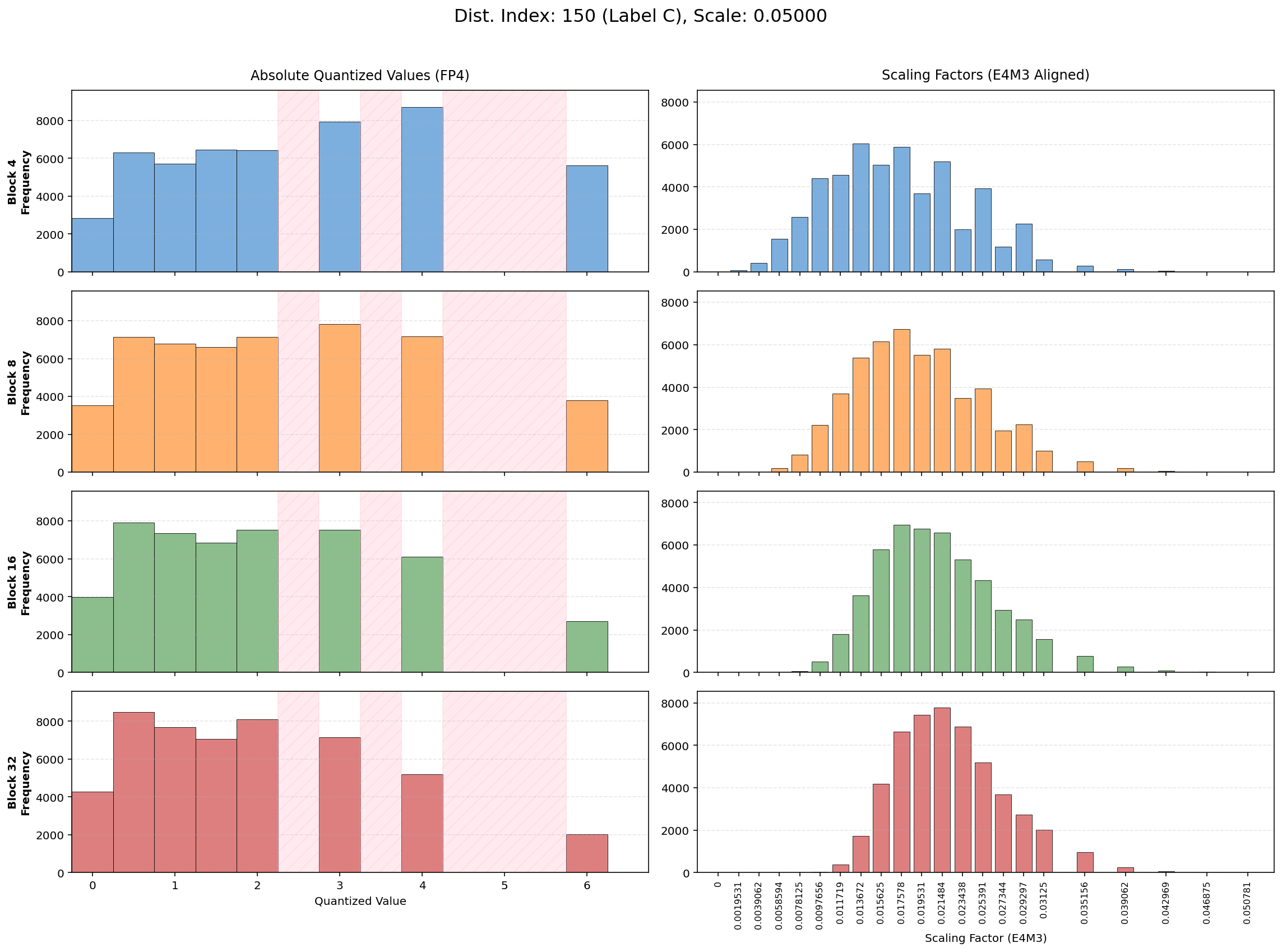}
        \caption{Region C}
        \label{fig:hist-brute-force-c}
    \end{subfigure}
    \vspace{-0.2cm}
    \caption{Brute-Force Search. \textbf{(a) Region A:} The scaling factors chosen align with the prevent-zero adjustment. Choosing a larger scaling value would result in underflowing too many entries \textbf{(b) Region B:} The scales show a somewhat wider distribution compared to 4-over-6, and significantly larger compared to abs-max. Although for smaller blocks a larger portion of entries fall into the magnitude 6 bin it is much less compared to  abs-max (comparable to 4-over-6). However, many more values fall into bins smaller than magnitude 4 compared to 4-over-6, suggesting in some cases the optimal result requires using only the lower portion of the FP4 range. \textbf{(c) Region C:} Visually comparable to the 4-over-6 case. Although similar to Region B, there are more entries in smaller magnitude bins.}
    \label{fig:hist-brute-force-combined}
\end{figure*}

\begin{figure*}[htbp]
    \centering
    \includegraphics[width=\textwidth]{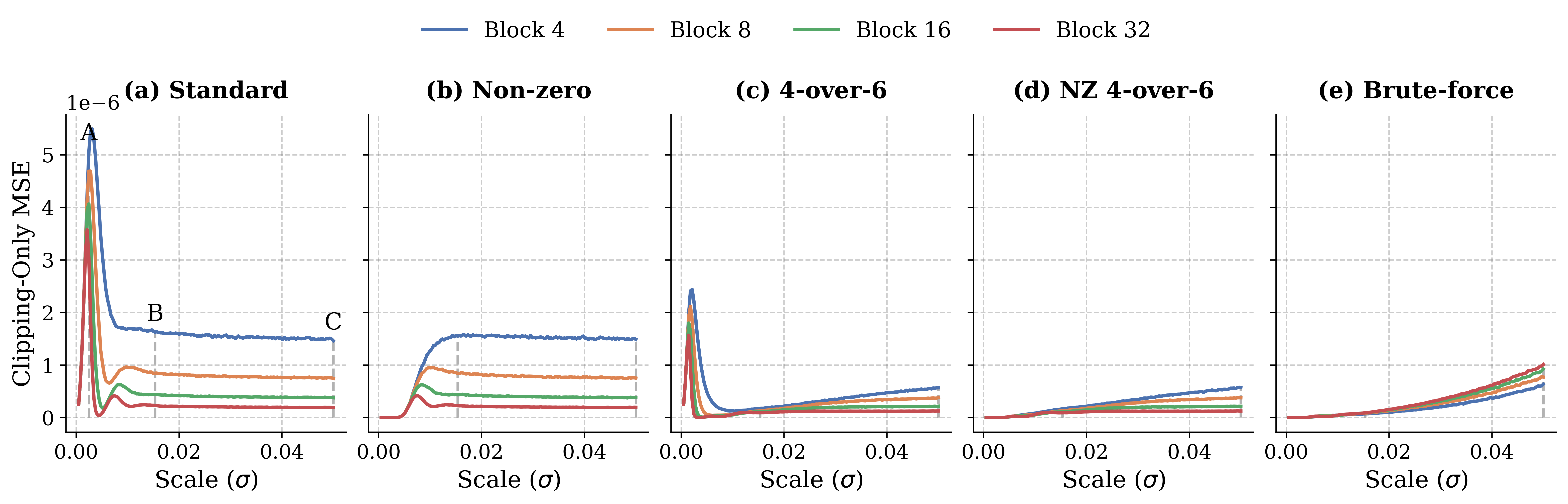}
    \vspace{-0.2cm} 
    \caption{Clipping only MSE. This plot is analogous to Figure~\ref{fig:mse_combined_horizontal}, showing only the clipping portion. Using 4-over-6 or brute force substantially reduces the clipping error for smaller block sizes, explaining much of the inversion of MSE vs block size at Region B.}
    \label{fig:mse_combined_clipping}
\end{figure*}

\subsection{A closer look at model weights}
\label{appendix:weights}

This this section we inspect the raw weight values of Granite-3.3-8B and Qwen2.5-14B shown in Figure~\ref{fig:final_figure} to better understand how quantization and the range of values impact the perplexity gap.

\begin{figure*}[htbp]
    \centering
    \begin{subfigure}[b]{0.24\textwidth}
        \centering
        \includegraphics[width=\textwidth]{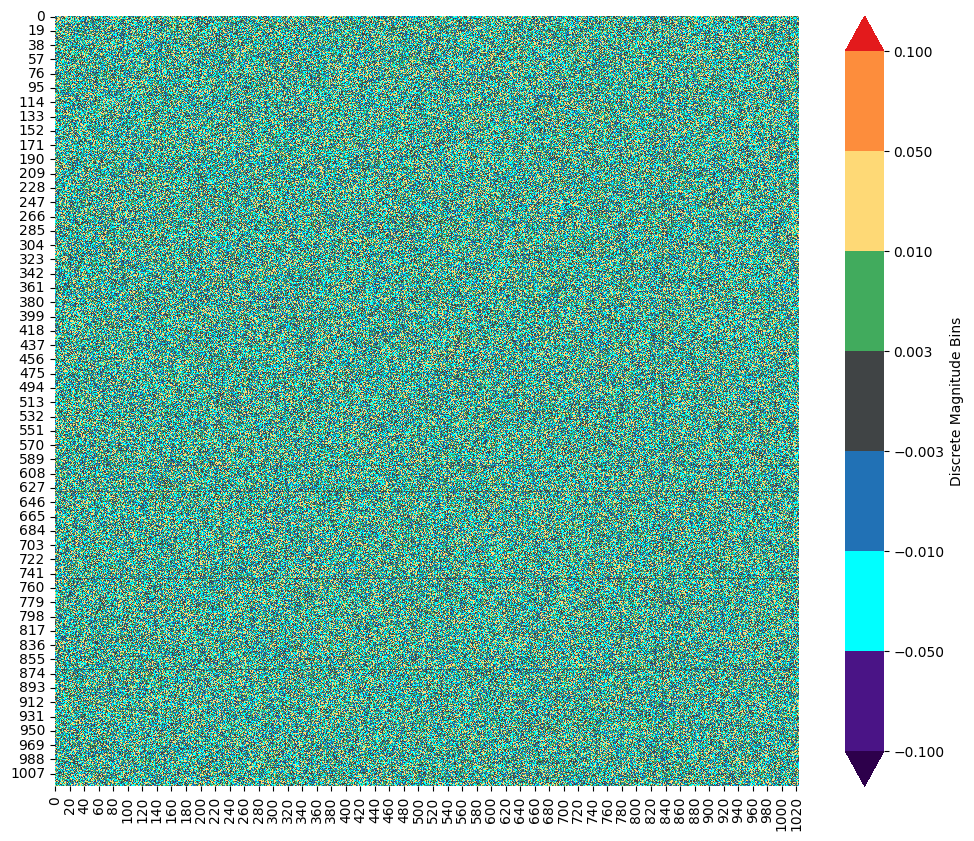}
        \caption{Granite weights for layer 2 down projection (sample).}
        \label{fig:granite-layer-2-hist}
    \end{subfigure}
    \hfill
    \begin{subfigure}[b]{0.24\textwidth}
        \centering
        \includegraphics[width=\textwidth]{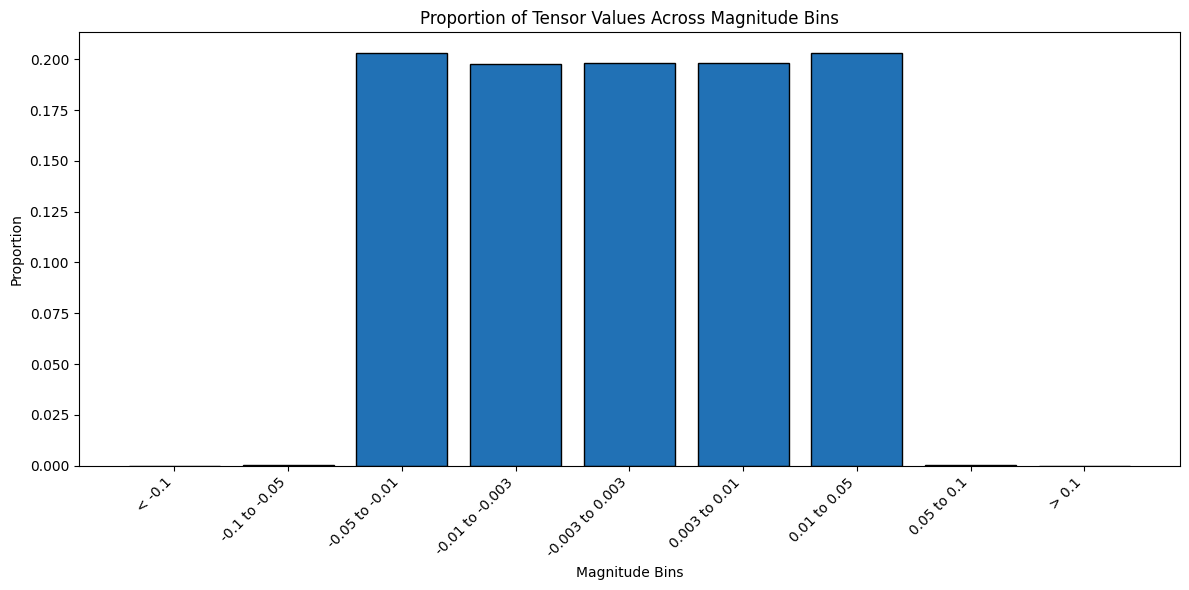}
        \caption{Granite weights for layer 2 down projection (distribution).}
        
    \end{subfigure}
    \hfill
    \begin{subfigure}[b]{0.24\textwidth}
        \centering
        \includegraphics[width=\textwidth]{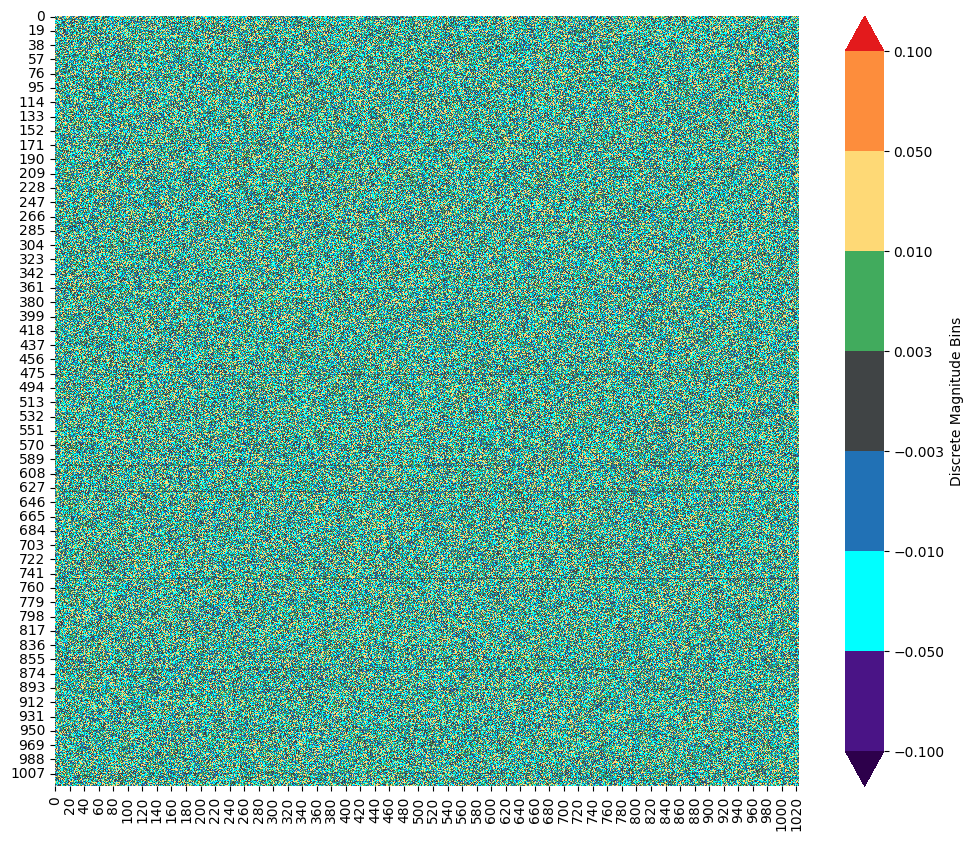}
        \caption{Granite weights for layer 2 down projection (sample).}
        
    \end{subfigure}
    \hfill
    \begin{subfigure}[b]{0.24\textwidth}
        \centering
        \includegraphics[width=\textwidth]{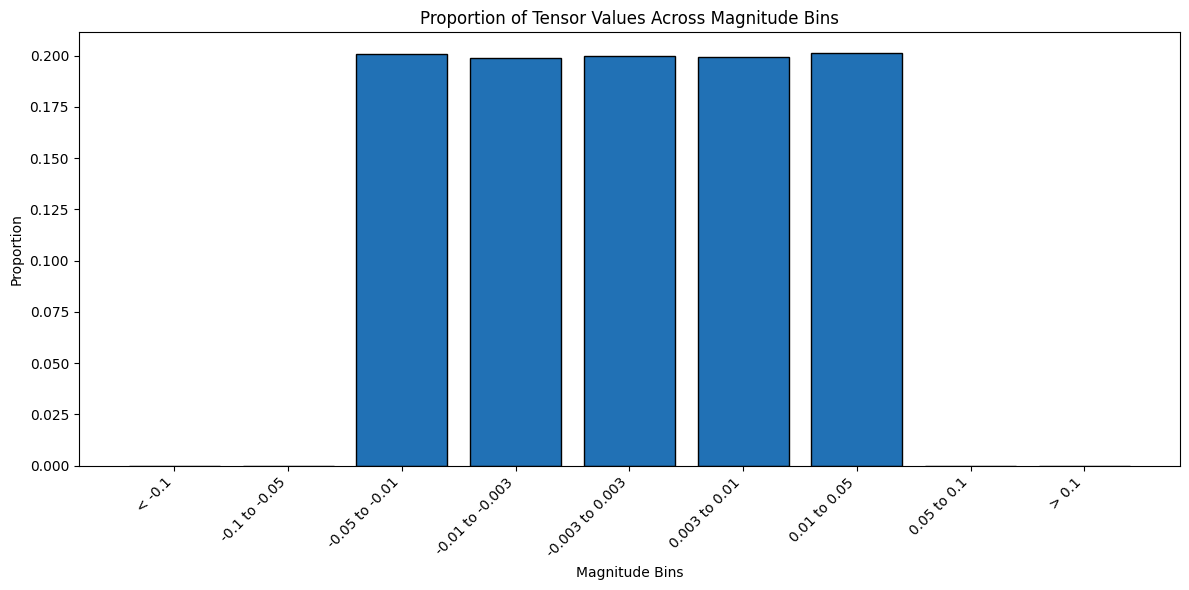}
        \caption{Granite weights for layer 25 down projection (distribution).}
        
    \end{subfigure}
    \vspace{-0.2cm}
    \caption{Granite 3.3 8B sample weights for layer 2 and layer 25 down projections. The distributions are similar at both parts of the network, with values approximately equally distributed among the five chosen bins.}
    \label{fig:gratine-weights}
\end{figure*}

\begin{figure*}[htbp]
    \centering
    \begin{subfigure}[b]{0.24\textwidth}
        \centering
        \includegraphics[width=\textwidth]{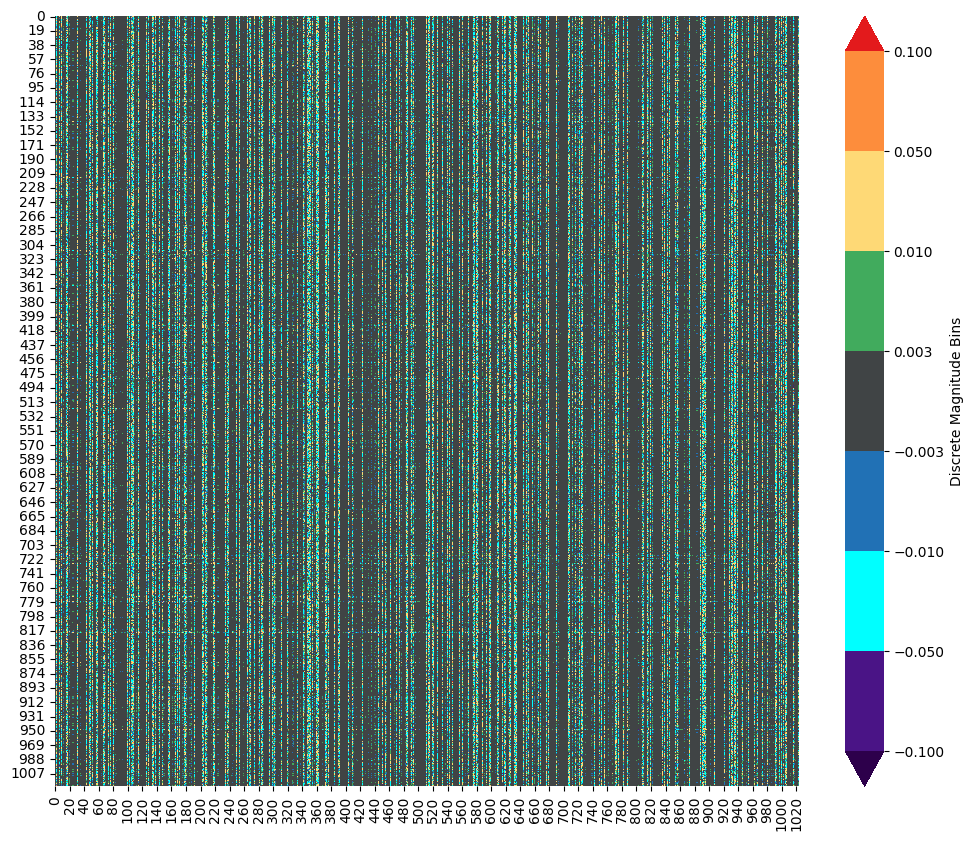}
        \caption{Qwen weights for layer 2 down projection (sample).}
        \label{fig:granite-layer-2-hist}
    \end{subfigure}
    \hfill
    \begin{subfigure}[b]{0.24\textwidth}
        \centering
        \includegraphics[width=\textwidth]{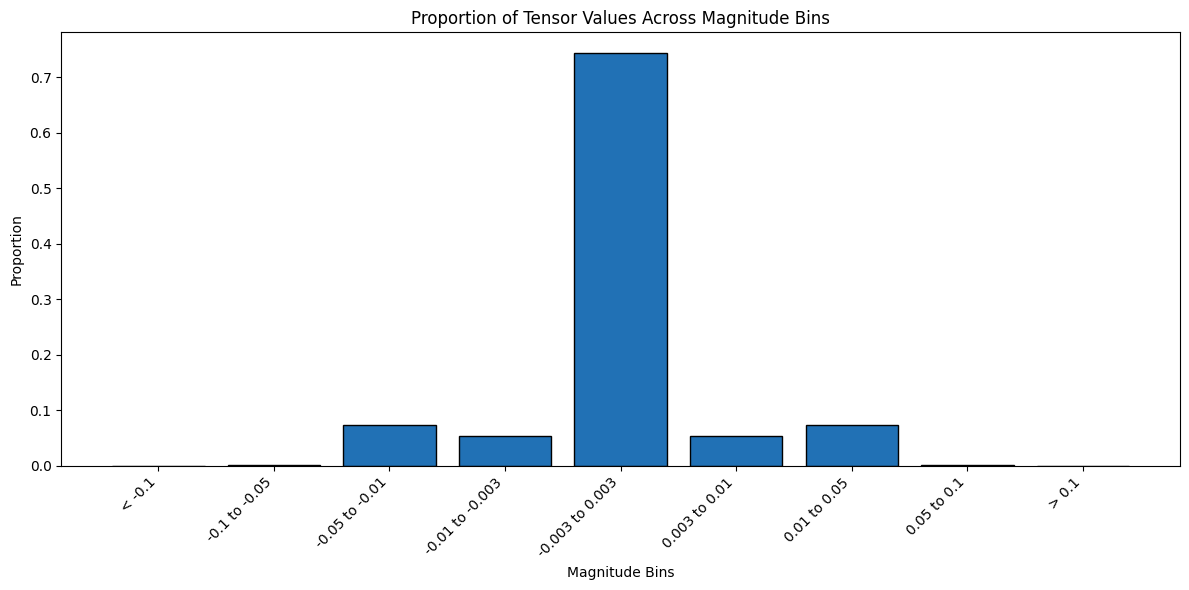}
        \caption{Qwen weights for layer 2 down projection (distribution).}
    \end{subfigure}
    \hfill
    \begin{subfigure}[b]{0.24\textwidth}
        \centering
        \includegraphics[width=\textwidth]{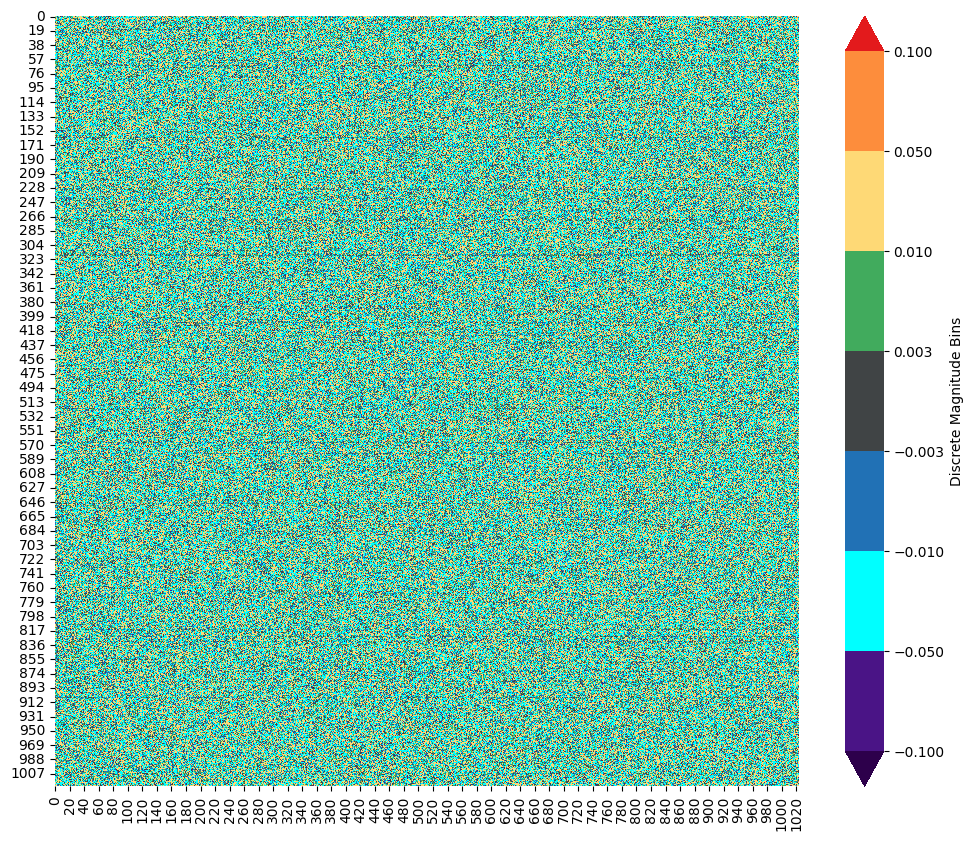}
        \caption{Qwen weights for layer 25 down projection (sample).}
    \end{subfigure}
    \hfill
    \begin{subfigure}[b]{0.24\textwidth}
        \centering
        \includegraphics[width=\textwidth]{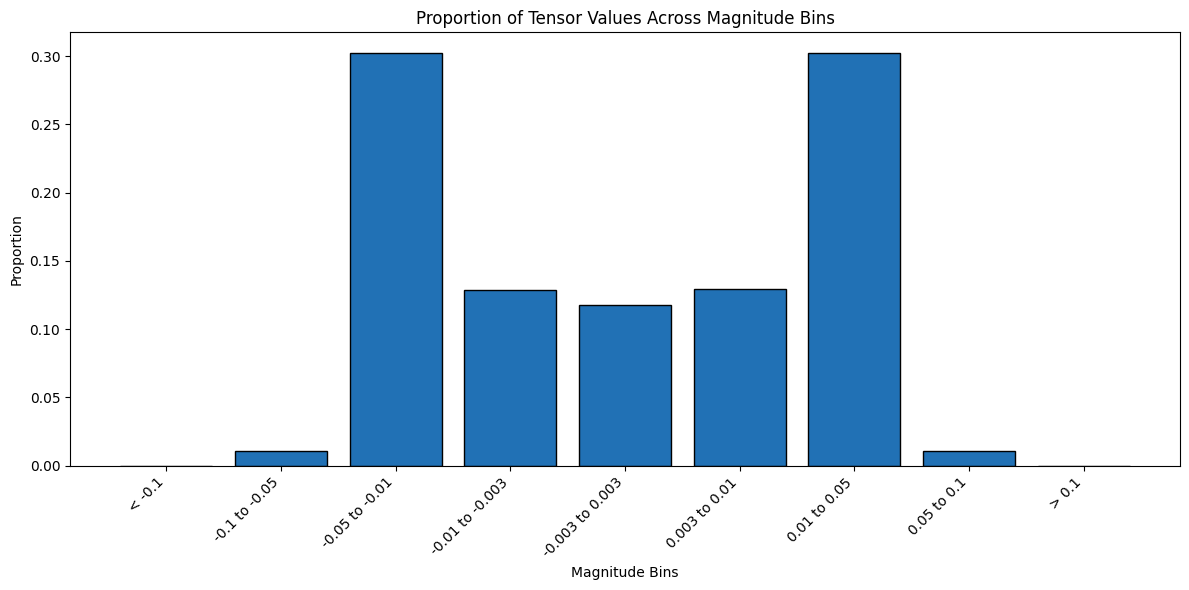}
        \caption{Qwen weights for layer 25 down projection (distribution).}
    \end{subfigure}
    \vspace{-0.2cm}
    \caption{Qwen 2.5 14B sample weights for layer 2 and layer 25 down projections. Earlier layers (a) show many columns with small values. In the aggregate the center bin [-0.003, 0.003] contains most of the values (b). However, the structural sparsity decreases for later layers, showing a more dense pattern (c). In the aggregate, larger magnitude bins contain most of the weight entries (d).}
    \label{fig:qwen-weights}
\end{figure*}

Figures~\ref{fig:gratine-weights} and~\ref{fig:qwen-weights} show the structure and aggregate distribution of the Granite and Qwen models examined. For Granite, the distribution was more consistent among different layers, while for Qwen we find individual columns of smaller values for earlier layers (suggesting effectively structural sparsity).
In later layers, the weight values of Qwen fall into larger magnitude bins at a higher proportion than Granite. 
This may help explain why despite the aggregate distributions in Figure~\ref{fig:final_figure} show more small values for Qwen, Granite was more sensitive to quantization with techniques requiring preservation of small values. In particular using E4M3 scaling factors without hierarchical scaling resulted in a consistently greater perplexity gap.

\begin{figure*}[htbp]
    \centering
    \includegraphics[width=\textwidth]{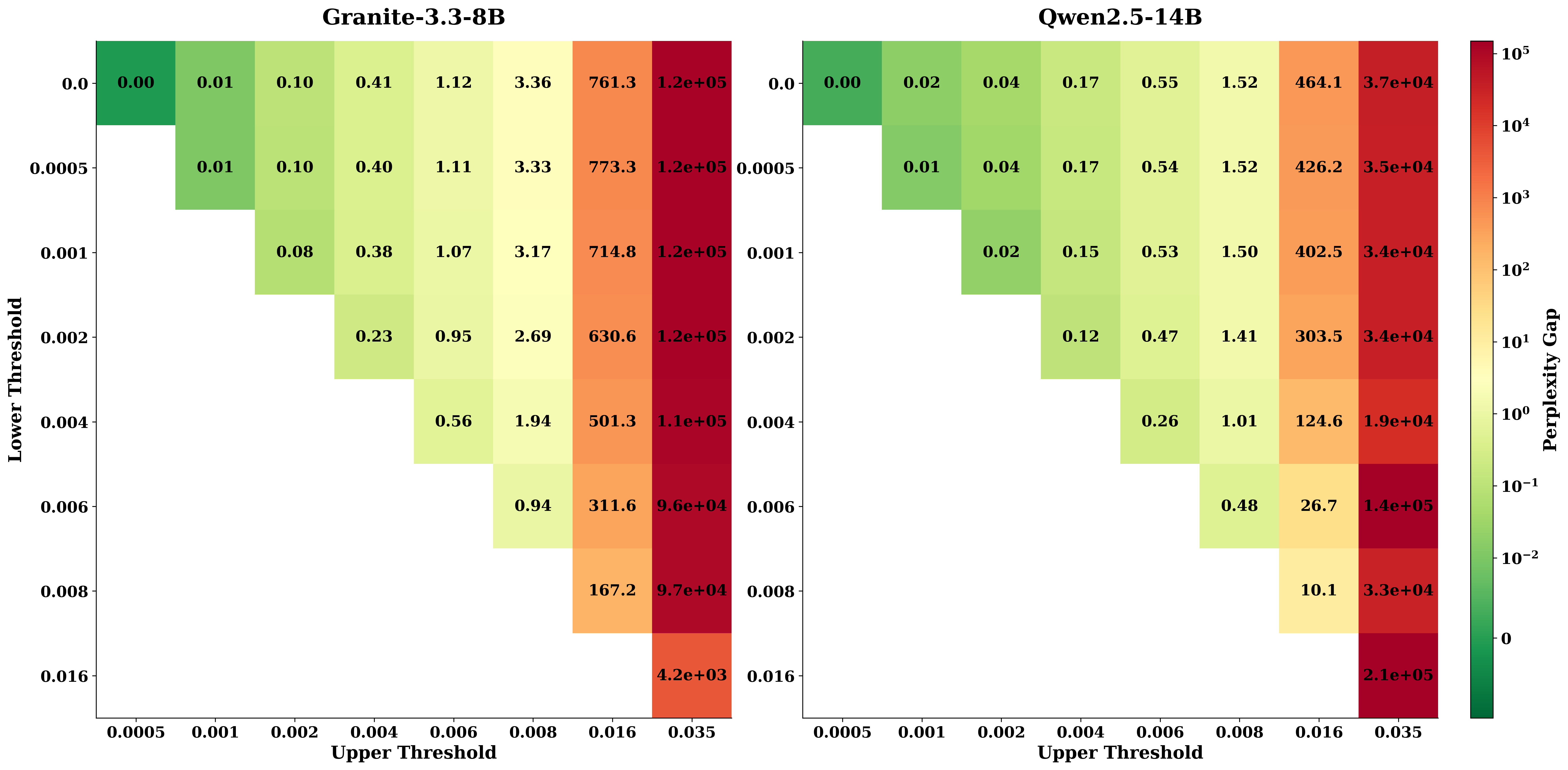}
    \caption{Unquantized ablation with mapping of specific entry value ranges to zero (between lower and upper thresholds specified on the y and x axis). The heatmaps illustrate the downstream perplexity gap when specific intervals of absolute weight and activation values are masked to zero at full precision. Sweeping boundaries from 0.0 up to 0.035 reveals general robustness of Qwen2.5-14B compared to Granite-3.3-8B. The diagonal which zeros out only one bucket shows that drastic increase in the perplexity gap comes later for the Qwen model, generally showing that larger values in activation and weights are more important for the Qwen model while Granite also relies on smaller values. This verifies our main findings that our prevent zero method is effective for Granite to recover some of the perplexity with small block sizes. Meanwhile, Qwen is not negatively affected by smaller block sizes, since its most impactful weights are larger than the region most impacted by the coarse E4M3 scaling factors. Additionally, this highlights that purely analyzing raw weight or activation values is not sufficient to pick good quantization schemes since Qwen had several layers with small values (see Fig.~\ref{fig:final_figure}), however, those value proved to be inconsequential for downstream perplexity.}
    \label{fig:zones_zero}
\end{figure*}

Figure~\ref{fig:zones_zero} shown an ablation where values between a lower and upper magnitude thresholds have been set to zero. This confirms that the smaller magnitude values seen in Qwen are less consequential.

\end{document}